\documentclass{article} 
\usepackage{collas2023_conference,times}
\usepackage{easyReview}

\usepackage{amsmath,amsfonts,bm}

\def\eqref#1{equation~\ref{#1}}

\DeclareMathAlphabet{\mathsfit}{\encodingdefault}{\sfdefault}{m}{sl}
\SetMathAlphabet{\mathsfit}{bold}{\encodingdefault}{\sfdefault}{bx}{n}
\usepackage{hyperref}
\hypersetup{
    colorlinks=true,
    linkcolor=red,
    filecolor=magenta,
    urlcolor=blue,
    citecolor=purple,
    pdftitle={Overleaf Example},
    pdfpagemode=FullScreen,
    }

\title{Towards Few-shot Coordination: Revisiting Ad-hoc teamplay challenge in the game of Hanabi}

\author{
Hadi Nekoei\thanks{Correspondence to: \texttt{nekoeihe@mila.quebec.}} \\
Mila, Université de Montréal \\
\And
Xutong Zhao \\
Mila, Polytechnique Montréal \\
\And
Janarthanan Rajendran \\
Mila, Université de Montréal\\
\AND 
Miao Liu  \\
IBM Research  \\
\And 
Sarath Chandar \\
Mila, Polytechnique Montréal \\
}

\collasfinalcopy 

\usepackage[utf8]{inputenc} 
\usepackage[T1]{fontenc}    
\usepackage{hyperref}       
\usepackage{url}            
\usepackage{booktabs}       
\usepackage{amsfonts}       
\usepackage{amsmath}
\usepackage{nicefrac}       
\usepackage{microtype}      
\usepackage{xcolor}         
\usepackage{mathtools}
\usepackage{microtype}
\usepackage{graphicx}
\usepackage{subfigure}
\usepackage{comment}
\usepackage{enumitem}

\usepackage{breqn}

\usepackage[algo2e]{algorithm2e}
\usepackage{algorithm, algorithmic}
\usepackage{tablefootnote}
\usepackage{xspace}
\usepackage{wrapfig}
\usepackage{caption}
\usepackage{amsthm}

\newcounter{example}[section]

\begin{document}

\maketitle
\begin{abstract}
Cooperative Multi-agent Reinforcement Learning (MARL) algorithms with Zero-Shot Coordination (ZSC) have gained significant attention in recent years. ZSC refers to the ability of agents to coordinate zero-shot (without additional interaction experience) with independently trained agents. While ZSC is crucial for cooperative MARL agents, it might not be possible for complex tasks and changing environments. Agents also need to adapt and improve their performance with minimal interaction with other agents. In this work, we show empirically that state-of-the-art ZSC algorithms have poor performance when paired with agents trained with different learning methods, and they require millions of interaction samples to adapt to these new partners. To investigate this issue, we formally defined a framework based on a popular cooperative multi-agent game called Hanabi to evaluate the adaptability of MARL methods. In particular, we created a diverse set of pre-trained agents and defined a new metric called adaptation regret that measures the agent's ability to efficiently adapt and improve its coordination performance when paired with some held-out pool of partners on top of its ZSC performance. After evaluating  several SOTA algorithms using our framework, 
our experiments reveal that naive Independent Q-Learning
(IQL) agents in most cases adapt as quickly as the SOTA ZSC algorithm Off-Belief Learning (OBL).
This finding raises an interesting research question: How to design MARL algorithms with high ZSC performance and capability of fast adaptation to unseen partners. As a first step, we studied the role of different hyper-parameters and design choices on the adaptability of current MARL algorithms. Our experiments show that two categories of hyper-parameters controlling the training data diversity and optimization process have a significant impact on the adaptability of Hanabi agents. We hope this initial analysis will inspire more work on designing both general and adaptive cooperative MARL algorithms.

\end{abstract}

\section{introduction}

Our everyday lives are filled with numerous cooperative multi-agent interactions. This includes our everyday regular activities such as crossing a traffic light, driving, buying and selling goods and services, activities at schools, offices and governments to name a few.
Artificial Intelligence (AI) agents and systems hold great promise in assisting and helping us with many of these day-to-day activities. It is of paramount importance for these AI agents to have the capabilities to coordinate with multiple humans and other agents (AI or otherwise). 
Building such  AI agents is challenging, as successful cooperation requires an agent to be able to predict and anticipate other agents' behaviors. Often agents have to do this with limited information of the state of world including that of the other agents, and also while other agents' actions being stochastic and changing over time.

Reinforcement Learning (RL) provides a general and scalable framework to model this challenging partially observable, non-stationary, multi-agent learning problem.
There has been a recent surge in interest within the AI research community toward building cooperative Multi-Agent RL (MARL) agents. In particular, the community has focused mainly on designing methods for the game of Hanabi \citep{bard2020hanabi} that enable trained RL agents to perform Zero-Shot Coordination (ZSC) without additional interaction with novel unseen agents~\citep{hu2020other, hu2021offbelief, lupu2021trajectory, cui2021klevel, nekoei2021continuous, lucas2022anyplay}. Hanabi, a partially-observable cooperative multi-agent benchmark, has been a particularly popular game in recent years to study MARL in a cooperative setting. However, most of these efforts are limited to the case of cooperating with novel agents trained independently but with the \textit{same underlying algorithm}. \citet{nekoei2021continuous} and \citet{lucas2022anyplay} are among the few approaches trying to evaluate agents in a more general scenario of cooperating with completely novel agents with maybe even different underlying learning algorithms.

While ZSC is an important and valuable feature that we would like our cooperative MARL agents to have, just focusing on ZSC is inherently restrictive.
In complex tasks and environments (such as in the real world), it may not be possible to learn everything relevant about the environment and the other unseen agents without ever interacting with them. It may not be possible to learn a universal coordination strategy that works well with all novel agents zero-shot.
Moreover, the world keeps changing. The environment, including the other agents in it and the task of interest, could change over time. 
We believe that along with the ability of ZSC, cooperative MARL agents should also have the ability to rapidly adapt with minimal interactions with other agents and improve their performance on top of their ZSC performance whenever possible.
The ability to adapt is quite important even after learning to best cooperate at any given moment in the agent's life~\citep{van2020loca}. This work aims to formally define adaptation to novel partners in the context of cooperative MARL and propose ways to measure it. 

Hanabi requires the agent to possess theory of mind to understand the intent of other agents and cooperate. \citet{bard2020hanabi} designed the Hanabi ad-hoc team play challenge to evaluate a Hanabi agent's ability to play and coordinate with a wide range of teammates the agent has never encountered before. 
\begin{quote}
    \textit{"Good strategies are not unique, and a robust player must learn to
    recognize intent in other agents’ actions and adapt to a wide range of possible strategies."}~\citep{bard2020hanabi}
\end{quote}
In particular, during the evaluation phase, the agent's performance is measured via the score achieved by the agent when it is paired with teammates chosen from a held-out pool of agents. However, the details of the ad-hoc teamplay challenge are left open for future work. Recently, there has been a transition in emphasis from addressing the ad-hoc coordination problem to addressing the specific challenge of creating algorithms that can coordinate with independently trained agents, while using the same high-level algorithm for training them individually also known as Label Free Coordination (LBF) problem~\citep{treutlein2021new}.
In this work, our goal is to bring back the focus to the ad-hoc coordination challenge which is a more general problem. In particular, we propose to extend this evaluation to measure not just the ZSC of a Hanabi agent, but also the Hanabi agent's ability to efficiently adapt and improve its coordination performance when paired with a held-out pool of agents on top of its ZSC performance.
For this, we define metrics such as the \textit{adaptation regret}, which measures the sum of the difference between the best-response score and the score achieved by the RL agent over time in the adaptation phase. Moreover, we investigate the choice of the partners which can greatly affects both zero-shot and adaptation performance of the learner. 

We carried out extensive experiments to fine-tune Hanabi agents using various pre-trained partners, and measured their adaptation regret, in order to better understand the current state-of-the-art (SOTA) ZSC algorithms' adaptation capabilities. We included various MARL algorithms from fully specialized Self-Play (SP) agents to the most generalist ZSC algorithms. It was not surprising to find that all of the SOTA algorithms needed millions of interactions with the partners to adapt well, as they lacked any mechanism to learn to adapt quickly during pre-training phase. However, we also discovered that naive Independent Q-Learning (IQL)\citep{tan1993multi} agents adapted to some of the partners as quickly as the SOTA Off-Belief Learning (OBL)~\citep{hu2021offbelief} algorithm, which is known to be excellent at ZSC. 
Therefore, a promising research direction would be to develop MARL algorithms that can perform both ZSC and Few-Shot Coordination (FSC), i.e., minimizing adaptation regret.

Finally, we investigated several hyper-parameters and architecture choices that potentially could influence the adaptability of the baselines. Our experiments show that two categories of hyperparameters (HPs) have a significant impact on the adaptation regret.  First, HPs that
affect data diversity such as the number of distributed threads and replay buffer size. The second type of HPs is the one
influencing the optimization process directly such as finetuning learning rate and batch size. We hope these initial investigations help to give some intuition to others who want to build both general and adaptive agents.

Our main contributions are summarized as follows
\begin{itemize}
    \item We conceptualize the few-shot coordination (FSC) setting for the ad-hoc teamplay challenge in multi-agent reinforcement learning. We accordingly propose the adaptation regret metric that evaluates how fast an agent adapts to unseen partners.
    \item We benchmark the adaptation performance of SOTA self-play and zero-shot coordination methods in the game of Hanabi. We discuss the inherent flaws of existing methods when paired with unseen partners.
    \item We study the effects of partners' diversity and hyper-parameters on the adaptation performance.
\end{itemize}
\section{related work}

The ad-hoc teamplay challenge in multi-agent reinforcement learning (MARL) requires agents to coordinate with unknown partners who are capable of contributing to the task. \citet{bowling2005coordination} and \citet{stone2000ad} were among the first to propose this challenge, while the authors of the Hanabi challenge~\citep{bard2020hanabi} explicitly propose it as a primary benchmark for future progress in cooperative MARL. However, ad-hoc teamplay challenge's details are left to be defined more precisely in the future work. 

Hanabi is a popular challenging cooperative game in which multiple different strategies are able to achieve strong performance. These strategies may not be able to coordinate well when paired with each other \citep{hu2020other,nekoei2021continuous}. 
Since Hanabi elevates the theory of mind reasoning about beliefs and intentions of other agents \citep{bard2020hanabi}, it becomes a highly suitable choice for studying ad-hoc coordination.
Other popular benchmark environments either do not exhibit the aforementioned characteristics (e.g., Google football \citep{kurach2020google}, most card games), have not been evaluated from a coordination perspective (e.g., Starcraft \citep{samvelyan2019starcraft}, MPE \citep{lowe2017multi}),
or are not commonly studied in the literature (e.g., small Hanabi with fewer colors/cards).
Nevertheless, the concept of few-shot coordination can be studied in other domains as well, as long as they require different strategies to adapt to novel partners.

Previous work on ad-hoc team play has involved learning diverse sets of policies and using Bayesian optimization, such as in \citet{canaan2019diverse} which uses the MAP-Elites algorithm~\citep{mouret2015illuminating} and an iterative Bayesian optimization approach \citep{brochu2010tutorial}. However, these approaches require meta-information and human knowledge that may not generalize to other problems. \citet{wu2021too} proposes Bayesian Delegation to perform efficient ad-hoc coordination by rapidly inferring the sub-tasks
of others.  Other related work includes MARL agents with access to observed behavior~\citep{barrett2017making, peysakhovich2017prosocial, lerer2019learning}, communication conventions between agents~\citep{sukhbaatar2016learning, mordatch2018emergence}, and zero-shot coordination (ZSC)~\citep{hu2020other, hu2021offbelief, nekoei2021continuous}.

In particular, the zero-shot coordination (ZSC) problem has received a surge of interest. In this problem, an algorithm is required to run independently and generate agents with high cross-play performance, i.e., the performance obtained by coordinating with other independently trained agents zero-shot, without any additional interaction with the new agents. To achieve this, methods such as leveraging symmetries of the problem~\citep{hu2020other}, training the best response to a diverse population of agents~\citep{nekoei2021continuous, lupu2021trajectory}, or making assumptions about prior actions~\citep{hu2021offbelief} have been proposed. In spite of the general definition of ZSC, the MARL community's focus has shifted towards designing agents that have good ZSC only with other agents also trained with the same underlying learning algorithm, but it does not impose any requirements on these agents to play well with unknown agents such as humans. Despite this, ZSC methods have yielded more versatile agents that have had limited success in transferring to the ad-hoc teamplay setting~\citep{hu2021offbelief}. 

More recently, \citet{zand2022fly} approaches the ad-hoc coordination challenge in the game of Hanabi by considering the problem of selecting a strategy from a finite set of previously trained agents using a posterior belief over the other agents' strategy, to play with an unknown partner. This approach requires having access to a pool of pre-trained policies and more importantly, a policy similar to the partner should be in the pool to get a good ad-hoc coordination performance. Lifelong Hanabi~\citep{nekoei2021continuous} and Anyplay~\citep{lucas2022anyplay} propose inter-crossplay evaluation of hanabi agents which is similar to ad-hoc teamplay challenge. \citet{nekoei2021continuous} also mentions briefly few-shot evaluation of agents to their partners while learning continually. However, none of these approaches consider a comprehensive evaluation of the adaptation capability of pre-trained Hanabi agents. In the context of the ad-hoc coordination in Hanabi, \citet{canaan2019diverse} also proposes Quality Diversity algorithms as a  class of algorithms to generate populations of diverse agents. These approaches can be complimentary to our benchmark to increase the diversity of our held-out pool of patners. 
\section{background}

\textbf{Dec-POMDPs:} In this paper we consider fully-cooperative Markov games. 
We model this setting with a Decentralized Partially-Observable Markov Decision Process (Dec-POMDP )~\citep{bernstein2002complexity, nair2003taming}, formally defined as a tuple $G = \{S, A, P, R, \Omega, O, N, \gamma\}$, with the set of states $S$, the set of actions $A$, the transition function $P$, the reward function $R$, the set of observations for each agent $\Omega$, the observation function $O$, the number of agents $N$, and $\gamma$ as the discount factor. The game is partially observable, with $o^i \sim O(o|i, s)$ as agent $i$’s observation of the global state, sampled from the (stochastic) observation function $O$. The game is also fully cooperative, thus agents share the same reward $r = R(s, \bm{a})$, conditioned on the joint action $\bm{a} = \left[ a^i \right]^N_{i=1}$ and the  global state $s$. At each timestep $t$, all agents are at the state $s_t$. Each agent has an action-observation history (\textit{AOH}) $\tau^i_t = \{o^i_0, a^i_0, r^i_0, \dots , o^i_t\}$, and selects action $a^i_t$ using a stochastic policy of the form $\pi^i_\theta(a^i|\tau^i_t)$. The transition function $P(s'|s_t, \bm{a}_t)$, conditioned on the joint action and the global state, transitions to the next state $s_{t+1}$. The goal is to maximize the expected return, $J = \mathbf{E}_\tau[R(\tau)]$, where $R(\tau) = \sum_t \gamma^t r_t$ is the discounted cumulative reward calculated using the discount factor $\gamma$.

\textbf{Deep MARL} has been successfully employed in various Dec-POMDP settings, as demonstrated by~\citet{oliehoek2016concise}. The conventional approach for employing deep Q-learning in Dec-POMDP settings is independent Q-learning (IQL)~\citep{tan1993multi}. In IQL, each agent treats other agents as part of the environment and learns an independent estimate of the expected return without incorporating the actions of other agents. For the sake of simplicity, this work uses the IQL setup with shared neural network weights $\theta$. We also pre-train agents with Value Decomposition Networks (VDN) algorithm~\citep{sunehag2017value} that learns a joint-action Q-function that consists
of the sum of per-agent Q-values to allow for off-policy learning in the multi-agent setting.

\textbf{R2D2:} In single-agent deep Q-learning~\citep{mnih2015human}, the agent predicts the anticipated total return for each action based on the state information. Given the partially observable setting, the techniques utilized in this study all utilize Recurrent Replay Distributed Deep Q-Networks (R2D2)~\citep{kapturowski2019recurrent} as their foundation. The Q function estimates action values based on the \textit{AOH} instead: $Q(\tau_t,a_t) = \mathbb{E}_{\tau} \left[ R_t(\tau_t) \right]$.
The R2D2 algorithm incorporates several modern best practices on top of deep Q-learning, including double-DQN~\citep{van2016deep}, dueling network architecture~\citep{wang2016dueling}, prioritized experience replay~\citep{schaul2016prioritized}, distributed training setup with parallel running environments~\citep{horgan2018distributed}, and recurrent neural network for dealing with partial observability.

\textbf{ZSC:} The prevalent method for learning in Dec-POMDPs is self-play (SP), which involves training RL agents with copies of themselves.  However, optimal policies learned through self-play often rely on arbitrary conventions that are agreed upon during training, which can be problematic in real-world scenarios where agents need to coordinate with other unknown AI agents and humans during testing. To address this issue, \citet{treutlein2021new} introduced the Zero-Shot Coordination (ZSC) setting, which requires learning algorithms that produce robust and unique solutions across multiple independent runs, and excludes arbitrary conventions as optimal solutions. Other-play (OP)~\citep{hu2020otherplay} is a method proposed to prevent agents from learning an arbitrary joint policy out of a set of equivalent but incompatible ones by enforcing equivariance to symmetries in the Dec-POMDP. On the other hand, \citet{hu2021offbelief} proposes Off-belief Learning (OBL) in the ZSC setting, which guarantees convergence to a unique policy by assuming prior actions were taken by a fixed random policy but future actions will be taken by the policy in training. This process can be iterated to train a new, higher-level policy using the trained policy from the previous iteration as the fixed policy. 



\section{Few-Shot Coordination}

We start by giving the reasons why we need to focus on \textit{Few-Shot Coordination (FSC)} besides Zero-Shot Coordination (ZSC) in section~\ref{sec:FSC_motivation}. Later, in section~\ref{sec:FSC_setup}, we formalize the FSC setting and explain adaptation regret as a metric to capture both ZSC and FSC capabilities of MARL algorithms.

\subsection{Motivation}\label{sec:FSC_motivation}

\begin{figure}[t!]
\centering
\begin{subfigure}[Cross-Play Matrix]{
    \centering
    \includegraphics[ width=0.4\linewidth]{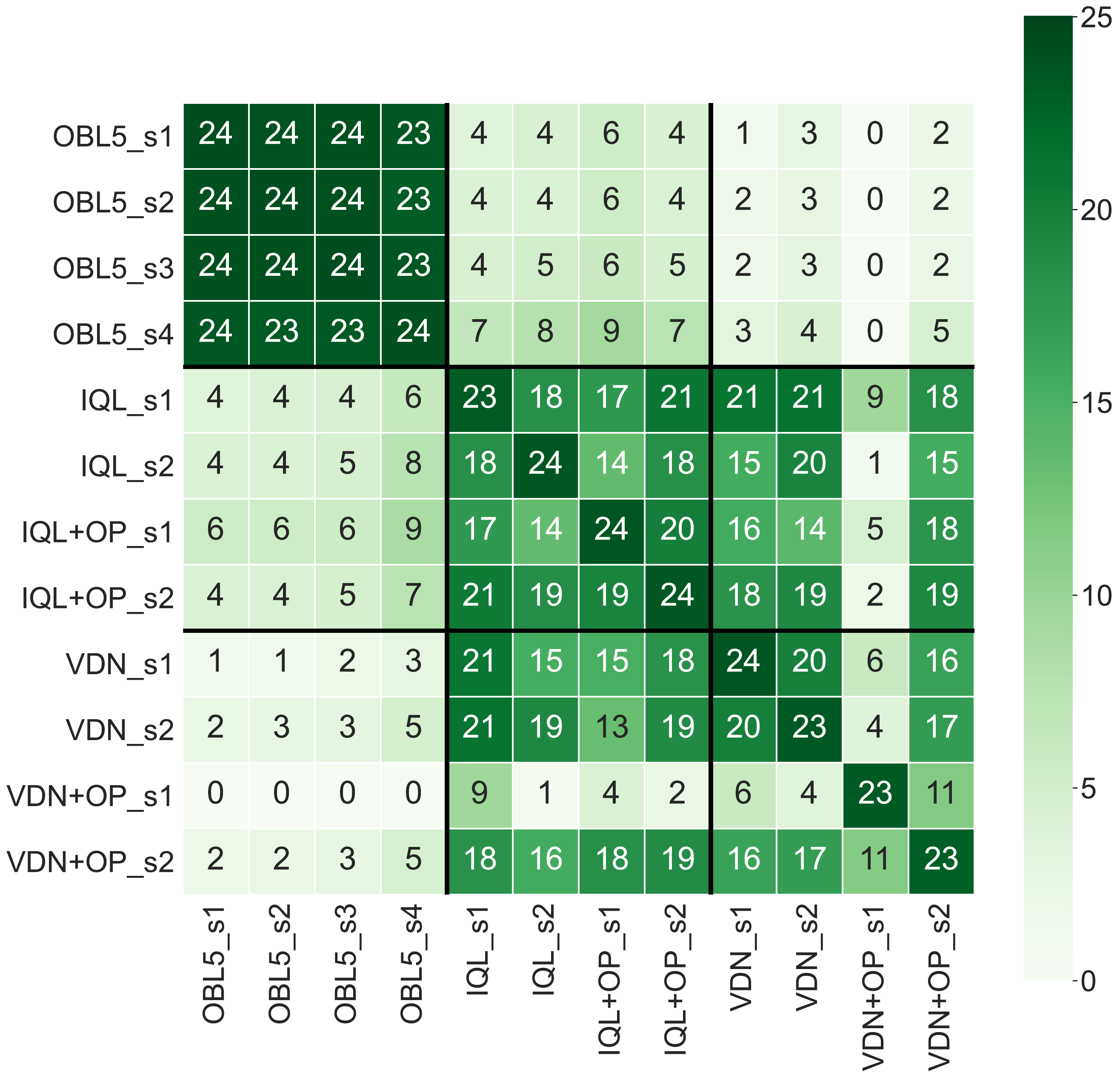}
    }
\end{subfigure}%
\begin{subfigure}[Adaptation curves]{
    \centering
    \includegraphics[ width=0.3\linewidth, height=0.3\linewidth]{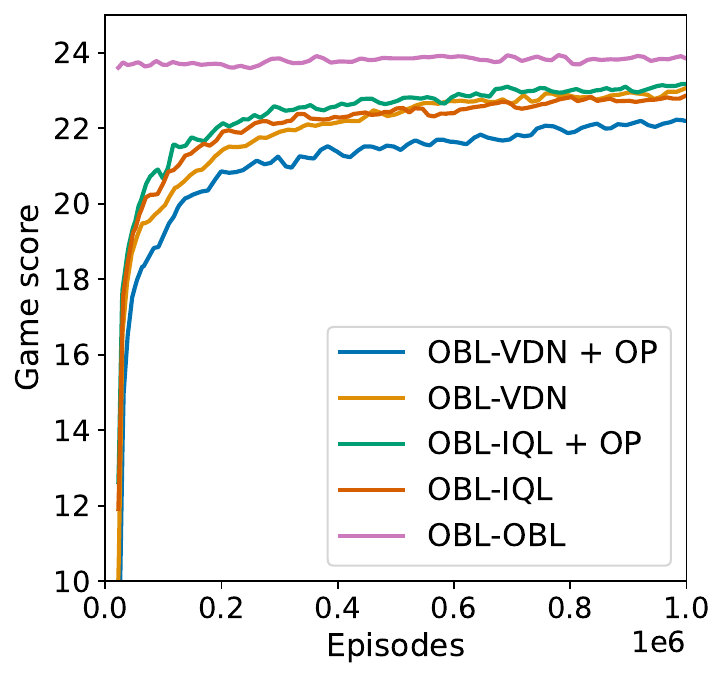}
    }
\end{subfigure}%
\caption{\em ZSC and adaptation performance of OBL algorithm with different self-play pre-trained agents. The maximum achievable score in Hanabi is 25. (a) Each row represents an agent with specific hyperparameters/architecture choices with high-level methods separated by black lines. Even though OBL algorithms perform well when paired with other agents trained with OBL, it performs poorly when paired with other SP agents like IQL or VDN. (b) The adaptation performance of an OBL agent when paired with different variants of IQL, VDN, and also an independent OBL agent. It is clear than the OBL agent requires millions of samples to adapt to these novel partners.}
\label{fig:why_few_shot} 
\end{figure}

The ad-hoc teamplay setting \citep{bard2020hanabi} aims to perform well when the agent is paired with \textit{any} other well-performing policy.
Despite its minimal assumptions, as pointed out by \citet{cui2021klevel}, ad-hoc teamplay in Hanabi fails in settings where there is little overlap between good SP policies and those that are suitable for coordination.
It can be unreasonable -- at least in the game of Hanabi -- to expect to have one single algorithm that can perform well with \textit{any} expert partner at test time as required by ad-hoc teamplay challenge.

The difficulty of coordinating with \textit{any} expert partner can be supported by several ZSC works in Hanabi.
In ZSC literature, the usual \textit{intra}-crossplay evaluation measures the zero-shot performance of two independent agents learned by the same underlying algorithms. 
However, previous work has shown a poor zero-shot performance of ZSC algorithms in the \textit{inter}-crossplay criterion, where the agent is paired with other agents pre-trained by different underlying algorithms \citep{nekoei2021continuous, lucas2022anyplay}. 
We also confirmed this phenomenon with more diverse agents as shown in Figure~\ref{fig:why_few_shot}(a). 
It is therefore sensible to work beyond zero-shot coordination and aim for fast adaptation towards those expert partners.

Our initial analysis shows that current SOTA MARL methods need an extremely large number of interactions with a new novel partner to adapt and improve their performance on top of their ZSC performance as shown in Figure~\ref{fig:why_few_shot}(b).
Recognizing the importance of the ability to adapt fast for AI agents, there have been several works recently on this topic in the single-agent RL setting, with a majority of them using techniques of meta-learning to learn to adapt fast. We hope this benchmark plays as a starting point to build such adaptive methods for the cooperative MARL setting.

Finally, excellent ZSC algorithms still are far from perfect in coordinating with humans~\citep{cui2021klevel, hu2021offbelief}. Moreover,
even though human players achieve the same performance with rule-based agents and ZSC learned agents, human players still strongly
prefer working with the rule-based agent~\citep{kim2020evaluation} viewing the Other-Play agent negatively, citing reasons such as a lack of bilateral understanding, trust, comfort, and perceived performance.
Even though there is no guarantee that FSC leads to an increase in subjective metrics without re-conducting the sentiment surveys, intuitively adaptive agents should become more similar to human strategies over time when paired with a human player.

Given these reasons, we believe that the community needs to focus on designing MARL algorithms that are able to adapt quickly with a few interactions to coordinate with AI agents or humans with special strategies while having a generalizable initial strategy capable of ZSC. Now we introduce Few-Shot Coordination (FSC) benchmark as a first step toward designing these algorithms.

\subsection{Benchmark Setup}\label{sec:FSC_setup}

\begin{wrapfigure}{r}
{0.4\textwidth}
\vspace{-10mm}
\centering
    \includegraphics[ width=0.85\linewidth]{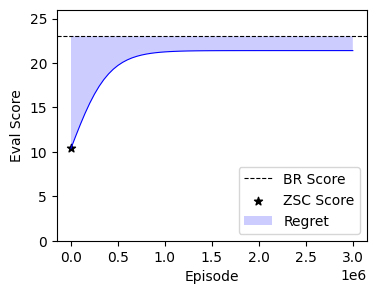}
\caption{\em Adaptation regret metric. This metric measures
the sum of the difference between the best-response score and the score achieved by the RL agent over time in the adaptation phase. Adaptation regret captures both ZSC and FSC performance. }\label{fig:regret_metric}
\end{wrapfigure}

Similar to most MARL settings, our benchmark consists of two phases: the training phase and the evaluation phase.
During the training phase, we do not pose any restrictions on the learner agent's learning algorithm, network architecture, or hyperparameter settings.
Any method can be used to train the learner agent, including SP, ZSC, or population-based methods.

During the evaluation phase, the objective is to evaluate how quickly the learner can adapt to a pool of unseen partner agents.
We define our evaluation metric, adaptation regret, for a learner $i$ and a fixed partner $j$ at evaluation episode $T$ as
\begin{align*}
    R_T (i, j) = T C^*_{j} - \mathbb{E} \left[ \sum^T_{t=1} C^t_{ij} \right],
\end{align*}
where $C^*_{j}$ and $C^t_{ij}$ denote the upper-bound performance for each partner and the performance of learner $i$ adapting to partner $j$ at time-step $t$, respectively. To aggregate the regret across different partners, we can take an average:
\begin{align*}
    R_T (i) = \texttt{aggr}(\{ R_T (i, j)\}_{j=1}^M)
\end{align*}

where $M$ is the number of partners. Instead of \texttt{aggr}, one can use standard averaging across partners or Inter-Quantile Mean (IQM) as suggested by \citet{agarwal2021deep} to make the metric robust to outliers.
We have several options for setting the upper-bound performance $C^*_{j}$.
In our current benchmark, we use the maximum achievable score in the game, which is 25.
Alternatively, $C^*_{j}$ could be the best-response (BR) score of partner $j$, which is the highest possible score that can be obtained by cooperating with partner $j$.
$C^*_{j}$ could also be the self-play score of the partner $j$.
It is worth noting that the choice of $C^*_{j}$ does not change the qualitative picture.
However, it can lead to large quantitative differences in the results.
For instance, if a partner $j$ is difficult to adapt to, using $C^*_{j}=25$ will make their contribution to the regret dominate the total regret.
The adaptation regret is illustrated in Figure~\ref{fig:regret_metric}.
It is worth noting that this metric captures both the ZSC and FSC performance of MARL methods.
For each agent, we evaluate its adaptation performance by its mean adaptation regret across a group of partners.

\begin{figure}[t!]
\centering
\begin{subfigure}[$D: 0.48, S: 0.94
$]{
    \centering
    \includegraphics[ width=0.3\linewidth]{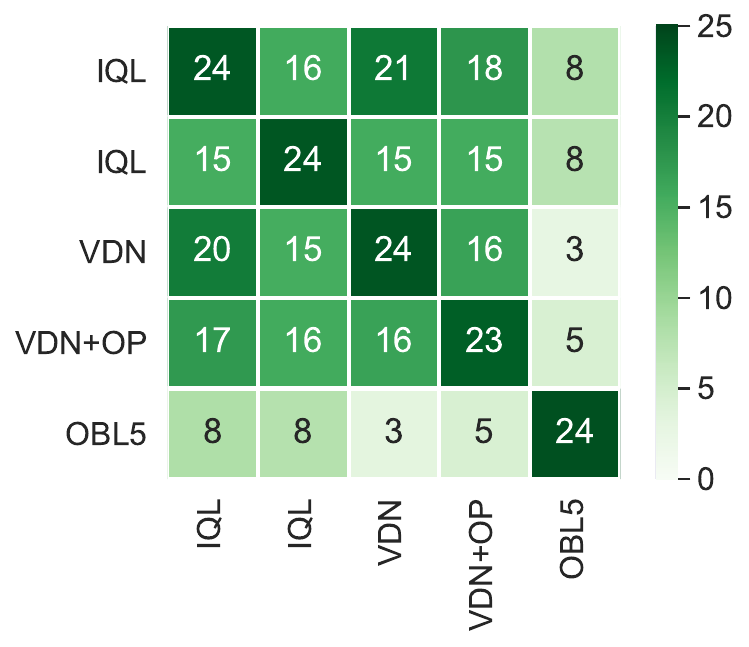}
    }
\end{subfigure}%
\begin{subfigure}[$D: 0.62, S: 0.94
$]{
    \centering
    \includegraphics[ width=0.3\linewidth]{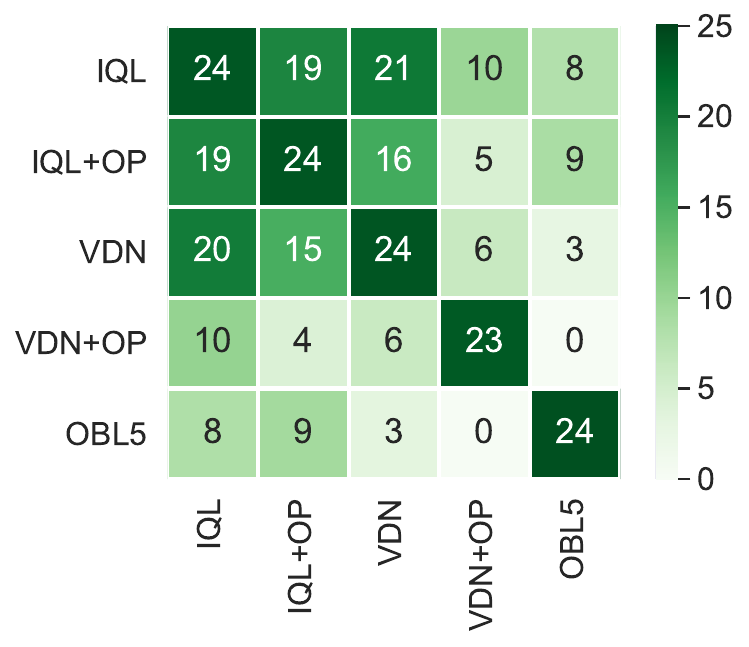}
    }
\end{subfigure}%
\begin{subfigure}[$D: 0.78, S: 0.96$]{
    \centering
    \includegraphics[ width=0.3\linewidth]{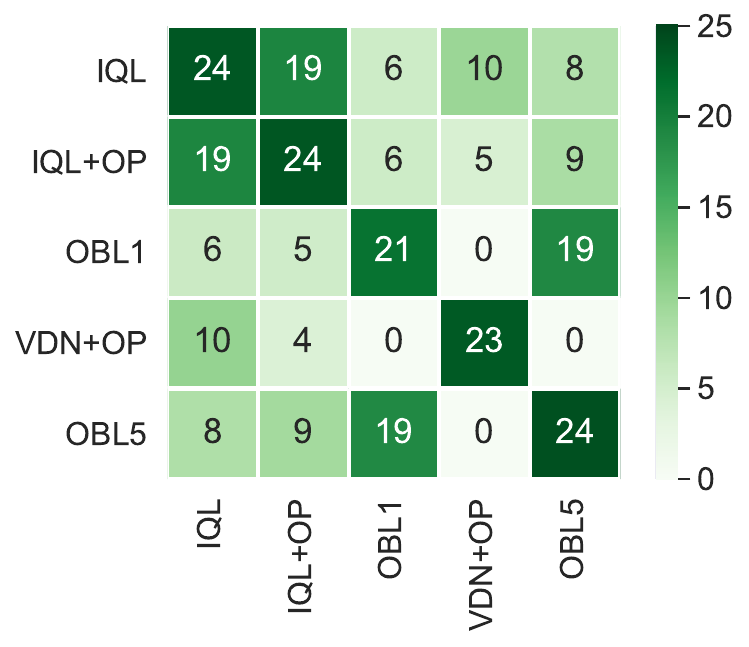}
    }
\end{subfigure}%
\caption{\em Three sets of pre-trained Hanabi agents with high strength (S) and different levels of diversity (D). The game score in each cell ranges from 0 to 25.}
\label{fig:partners} 
\end{figure}

As the adaptation performance highly depends on the set of partners,
we assume each partner meets two basic requirements.
First, each partner agent should demonstrate its capability of achieving strong cooperation with some other agent.
For instance, a SP agent can reach a high score with itself, or a ZSC agent can reach a high ZSC score with another agent optimized with the same ZSC algorithm.
This states that each partner $j$ should have a high $C^*_{j}$ value.
We define $S_j = C^*_{j} /25 \in [0,1]$ to represent the strength of individual partner $j$, and $S = \frac{1}{n} \sum_j S_j$ to indicate the strength of a set of partners, where $n$ is the number of partners.
This requirement is important as a random partner is less meaningful to evaluate the learner's adaptability.
Second, in order to ensure the learner does not overfit to some arbitrarily specialized partners, the set of partners should have diverse playing strategies.
Given that a set of partners have high strength $S$, intuitively they are diverse if at the same time they have low cross-play scores with each other.
We define a soft metric to represent the diversity level of partners:
\begin{align*}
    D =  1 - \frac{1}{n^2-n}\sum_{i\neq j} \frac{C_{ij}}{25},
\end{align*}
where $C_{ij}$ is the cross-play score of partner $i$ and partner $j$ for a given group of partners.
Figure~\ref{fig:partners} shows three example sets of partners with different levels of diversity.
Note that in the case of partners with fully diverse strategies, $D = 1$ and in the case of partners with fully aligned strategies, $D = 0$.


According to these two requirements, each partner could be an agent trained to optimize for SP or ZSC performance.
It could also be an offline-learning agent trained with human data or a rule-based agent with hardcoded strategies.
We assume each partner remains fixed throughout the evaluation. This assumption is reasonable for the FSC setting in which the goal is to adapt to new partners in a few episodes. Nonetheless, extending our benchmark to include a learning partner is an intriguing avenue that we will leave for future research. 


\section{Experiments}
\label{sec:experiments}

In this section, we report the benchmarking results of several MARL algorithms following the setting described in section~\ref{sec:FSC_setup}. 

First, we gathered a pool of diverse pre-trained agents including self-play agents such as IQL and VDN and agents trained to perform well in terms of ZSC such as OBL, IQL+OP, and VDN+OP. Then for each method we chose randomly an agent as a learner. Finally, based on the aforementioned requirements in section~\ref{sec:FSC_setup}, we choose the partners as shown in Figure~\ref{fig:partners} to evaluate adaptation.
More details on the agents, their network architecture, etc are described in Table~\ref{tab:exact-architectures} in Appendix~\ref{sec:experiment-setup}\footnote{All the pre-trained agents and the evaluation code are available at \href{https://github.com/chandar-lab/adaptive-hanabi.git}{https://github.com/chandar-lab/adaptive-hanabi.git}.}.
During adaptation, each partner's policy remains fixed.
Moreover, regardless of the training algorithm used to obtain the learner, each learner performs gradient updates as an IQL agent using the same hyperparameter setting as in the pre-training phase.
Each learner’s adaptation performance against each partner is computed by averaging across 1000 independent games (i.e., seeds). Then for each learner, its adaptation performance is aggregated by averaging the adaptation performance across 5 partners.



\subsection{Benchmark results}\label{sec:benchmark_results}

\begin{figure}[t!]
\centering
\begin{subfigure}[Total adaptation regret]{
    \centering
    \includegraphics[ width=0.32\linewidth]{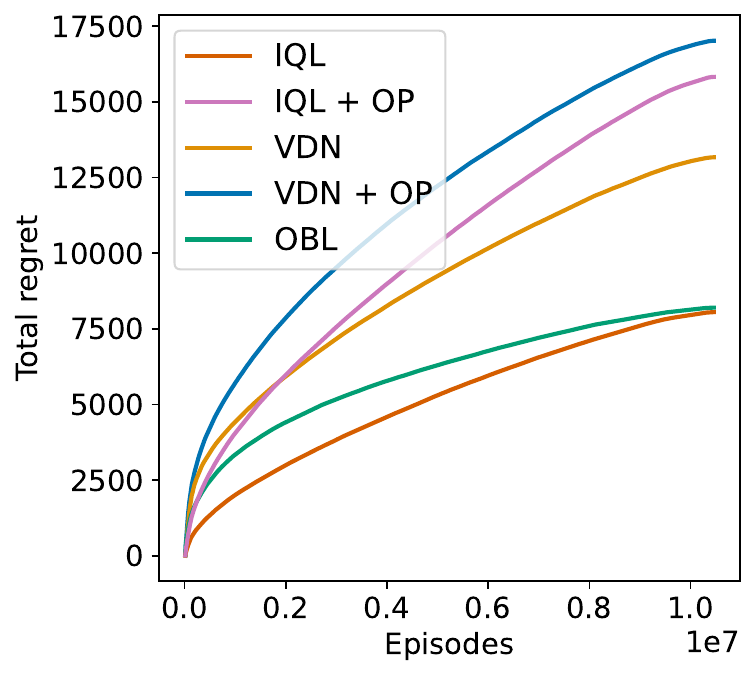}
    }
\end{subfigure}%
\begin{subfigure}[Average adaptation regret]{
    \centering
    \includegraphics[ width=0.3\linewidth]{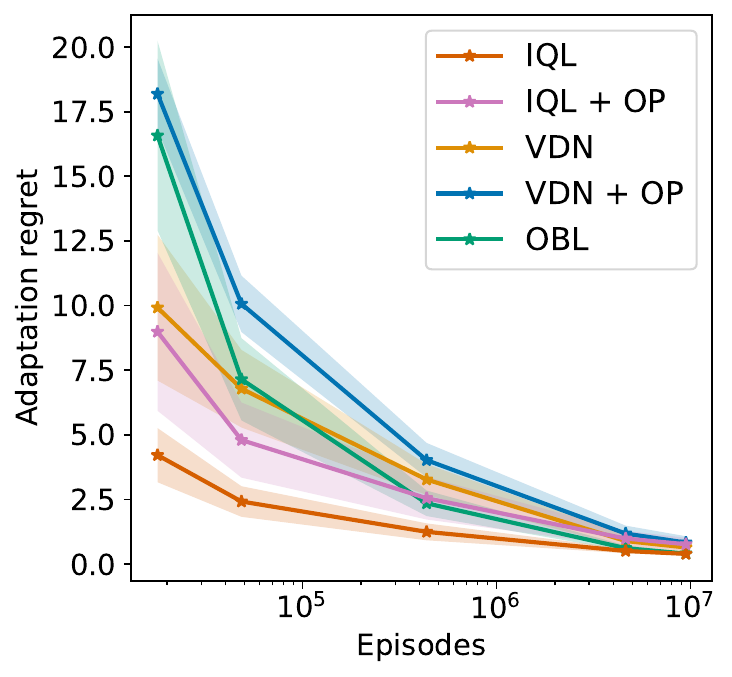}
    }
\end{subfigure}%
\begin{subfigure}[Game score]{
    \centering
    \includegraphics[ width=0.3\linewidth]{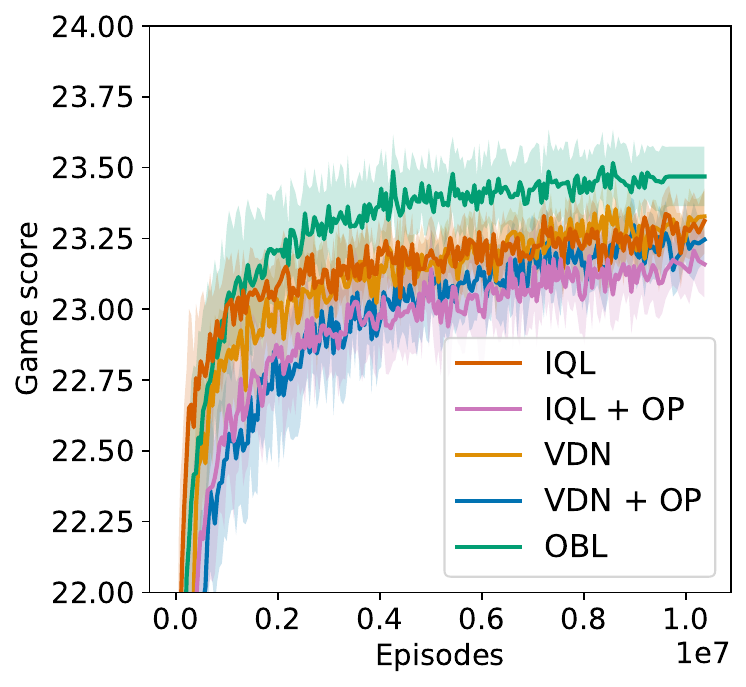}
    }
\end{subfigure}%
\caption{\em Results on adapting 5 different agents to partners from Figure~\ref{fig:partners}(b). Refer to Figure~\ref{app_fig:benchmark_results} in the appendix for the adaptation results of the other two sets of partners from Figure~\ref{fig:partners}. (a) Total regret shows that IQL has lower regret initially than OBL but as OBL finds the best response to the partners, it grows slower than IQL (b) Average adaptation regret over past episodes (c) Game score average across the partners. If zoomed in, IQL adapts faster initially but it converges to a lower final performance. The shaded area is the standard error across 5 different partners.}
\label{fig:benchmark_results} 
\end{figure}

Benchmarking results are summarized in Figure~\ref{fig:benchmark_results}, where each curve represents the mean performance of each learner across five partners.
The left figure shows the cumulative adaptation regret of each learner over evaluation episodes, the middle figure shows the adaptation regret averaged by the episode number at different points of evaluation, and the right figure shows the actual game score obtained. 
In Appendix~\ref{sec:additional-exp} we provide additional experiments on agents with different levels of strength and diversity.



\begin{figure}[h!]
\centering
\begin{subfigure}[Partners' Diversity = 0.48]{
    \centering
    \includegraphics[ width=0.3\linewidth]{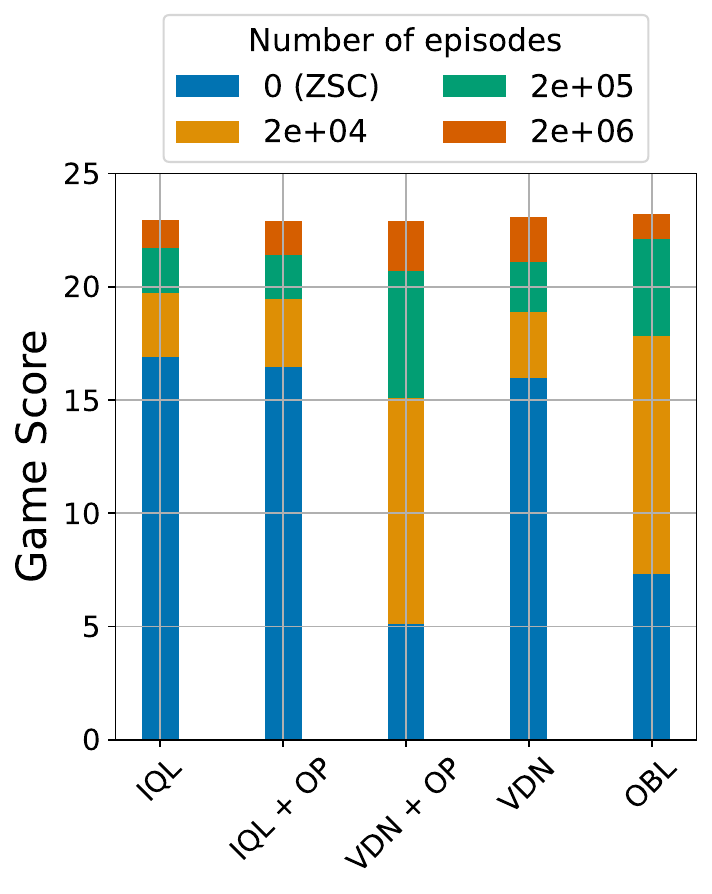}
    }
\end{subfigure}%
\begin{subfigure}[Partners' Diversity = 0.62]{
    \centering
    \includegraphics[ width=0.3\linewidth]{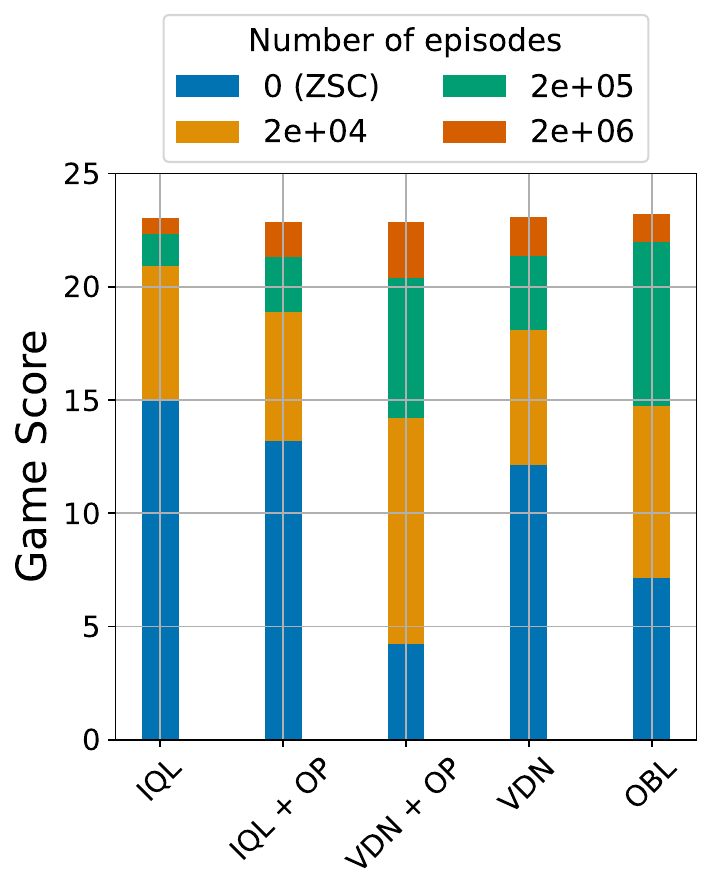}
    }
\end{subfigure}%
\begin{subfigure}[Partners' Diversity = 0.78]{
    \centering
    \includegraphics[ width=0.3\linewidth]{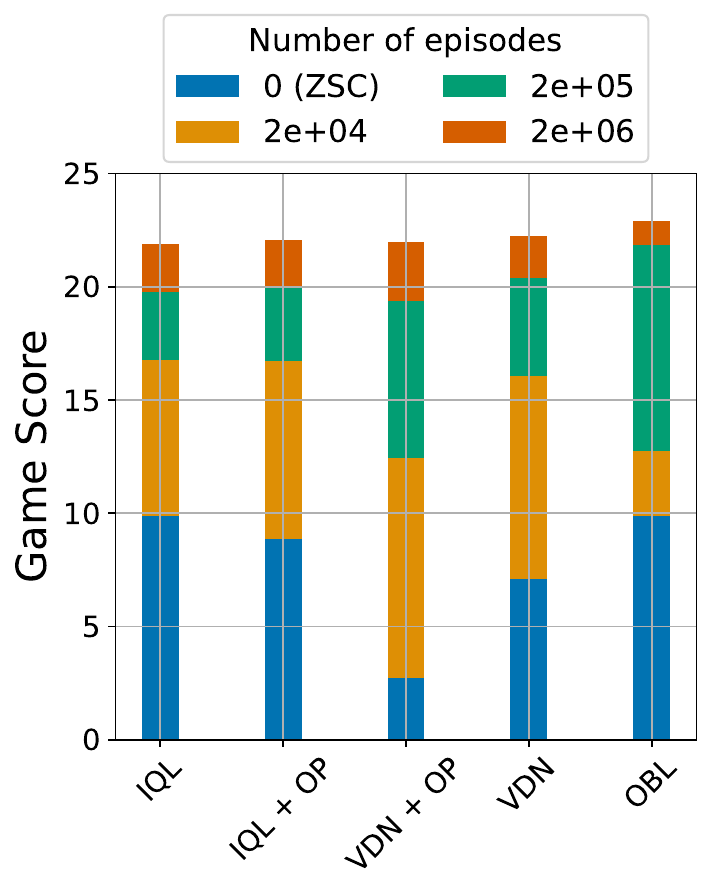}
    }
\end{subfigure}%
\caption{\em Each color represents the game score of a learner adapting to the partners from Figure~\ref{fig:partners} after different number of episodes denoted by the legends. These plots can be complementary to adaptation regret plots with ZSC performance included explicitly. The \texttt{aggr} used for this plots is IQM.}
\label{fig:barplot_game_scoer} 
\end{figure}

In general the results demonstrate the lack of efficient adaptability of commonly used methods.
To adapt to a partner independently trained with a different algorithm or architecture, they require millions of episodes, which is several orders of magnitude higher than the amount of data needed for few-shot learning in supervised learning.

The SOTA ZSC algorithm OBL achieves a higher game score among the different methods eventually.
Since the partners also include an OBL agent, as a ZSC method it is normal to reach a relatively higher final score.
However, OBL has a higher average regret than IQL and VDN at least initially, despite its high performance in the ZSC problem where both agents are trained with OBL.
This result highlights the significant impact of the choices of partners on performance. 
Even though the adaptation regret aims to capture both ZSC and FSC performance, looking solely at the adaptation regret sometimes can be misleading as it can be sometimes dominated by either very small or very large ZSC performance. Therefore, we also visualize the game score at different timesteps during adaptation that we care about in Figure~\ref{fig:barplot_game_scoer} for all three sets of partners with different diversity levels. When examining Figure~\ref{fig:barplot_game_scoer}(a) and Figure~\ref{fig:barplot_game_scoer}(b), we observe that although OBL has achieved a lower game score after $t = 2e4$ adaptation steps compared to other methods, we should note that it has started from a much lower zero-shot performance at $t = 0$ as indicated by the blue color. On the other hand, in the case of high diversity partners shown in Figure~\ref{fig:barplot_game_scoer}(c), both IQL and OBL start from the same ZSC performance. However, IQL adapts faster and achieves better performance at $t = 2e4$ but it is eventually outperformed by OBL after  $t = 2e5$. Therefore, Figure~\ref{fig:barplot_game_scoer} alongside Figure~\ref{fig:benchmark_results} provides us with a better understanding of the adaptation performance of various methods.
Strong ZSC performance does not suffice to guarantee good results when the learner is paired with other methods in the FSC setting.
Another interesting observation is that both IQL and VDN reach a better performance than their other-play counterpart.
This observation additionally demonstrates that the direct application of successful ZSC techniques may not lead to performance improvement in few-shot adaptation.



We performed an ablation study on the choice of the upper-bound performance $C_j^*$ and found that it has a negligible impact on the average adaptation regret. (Refer to appendix~\ref{app:upper_bound}). Nevertheless, if the SP score of each partner is employed as $C_j^*$, the differences between algorithms become more noticeable regarding the total regret. This is due to the fact that the average regret is already normalized, and modifying the upper-bound score only slightly shifts the regret curves. Nonetheless, this small shift is reflected more prominently in the total regret curve.


\subsection{The role of hyper-parameters in adaptation}\label{sec:HPS_results}

\begin{figure}[t!]
\centering
\begin{subfigure}[Average adaptation regret, $\downarrow$ better]{
    \centering
    \includegraphics[ width=1\linewidth]{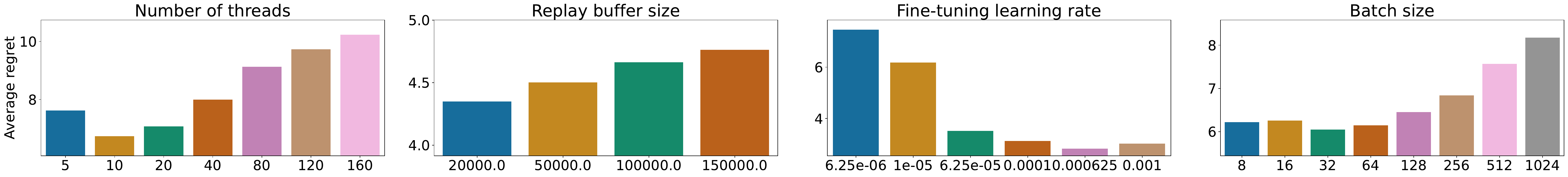}
    }
\end{subfigure}%
\begin{subfigure}[Perfect score, $\uparrow$ better]{
    \centering
    \includegraphics[ width=1\linewidth]{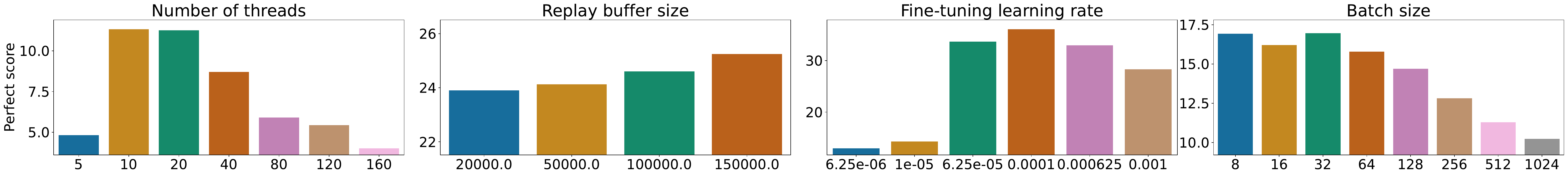}
    }
\end{subfigure}%
\caption{\em The role of different hyperparameters on adaptability and performance of an IQL agent adapting an OBL agent. (a) Average adaptation regret that is total adaptation regret divided by the number of episodes (b) The percentage of the games finished with a perfect score of 25. All of the HPs show a significant influence on both adaptability and the perfect score. }
\label{fig:iql_to_obl_hps} 
\end{figure}

\begin{figure}[h!]
\centering
\begin{subfigure}[Average adaptation  regret, $\downarrow$ better]{
    \centering
    \includegraphics[ width=1\linewidth]{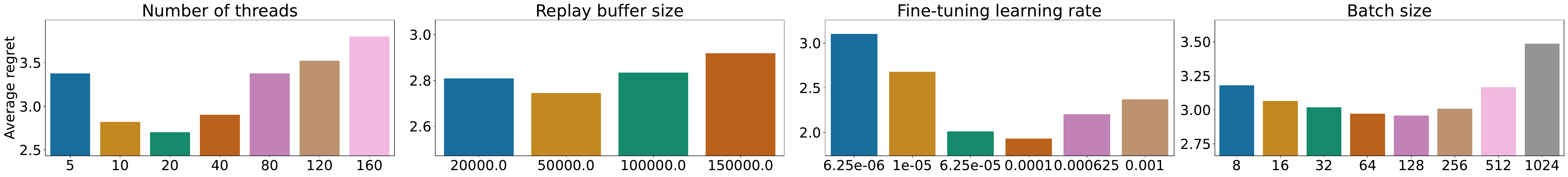}
    }
\end{subfigure}%
\begin{subfigure}[Perfect score,  $\uparrow$ better]{
    \centering
    \includegraphics[ width=1\linewidth]{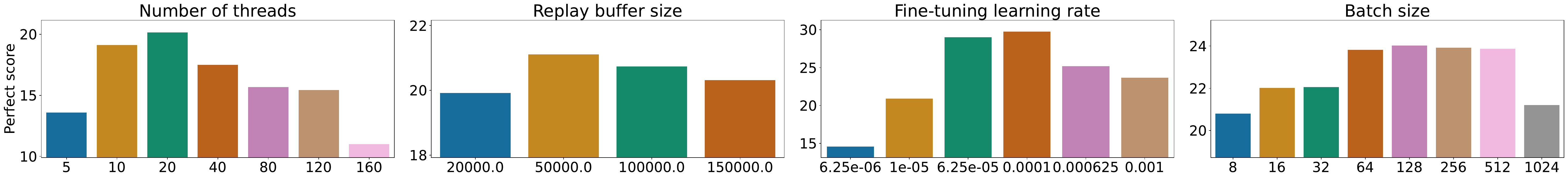}
    }
\end{subfigure}%
\caption{\em The role of different hyperparameters on adaptability and performance of an OBL agent adapting to an IQL agent.  (a) Average adaptation regret (b) The percentage of the games finished with a perfect score of 25. All of these HPs show a significant influence on both adaptability and the perfect score.}
\label{fig:obl_to_iql_hps} 
\end{figure}

We consider two primary categories of hyper-parameters (HPs) that can impact the adaptation regret. The first category includes HPs that affect data diversity, such as the number of distributed threads and replay buffer size. The second category includes HPs that directly influence the optimization process, such as finetuning learning rate and batch size. We performed the hyper-parameters tuning around the original values used in~\cite{hu2021offbelief} and reported the results in Figures \ref{fig:iql_to_obl_hps} and \ref{fig:obl_to_iql_hps}. In particular, the effect of the aforementioned HPs on both (a) adaptation regret and (b) the perfect score is studied. While the former measures how sensitive is the adaptability w.r.t these HPs, the latter shows what percentage of the evaluation games are finished with a perfect 25 score for different values of each HP.

\textbf{Number of threads:} \texttt{num_threads} and \texttt{num_games_per_thread} are two HPs that their multiplication determines the number of parallel independent games running and generating game episodes. 
A lower number of threads and number of games per thread mean more policy updates per environment step. The original values used by \citet{hu2021offbelief} are \texttt{num_theads}=$80$ and \texttt{num_games per_thread}=$80$. As we decrease the values of these two HPs, the adaptation improves (20 seems to be the best) but if it’s too small (10 or 5) it starts to have an adverse effect which can be because of the replay buffer having less diverse experience.

\textbf{Replay buffer size:} \texttt{replay_buffer_size} determines the maximum number of episodes stored in the buffer. The original value is $1e5$. 
From Figure~\ref{fig:iql_to_obl_hps}, smaller buffer sizes seem to aid adaptation initially, but overly small sizes may reduce data diversity and result in poor performance. The best value is around $5e4$. In Figure~\ref{fig:obl_to_iql_hps} however, the smallest buffer size tested seems to perform best.

\textbf{Fine-tuning learning rate:} The optimizer used throughout the experiments is Adam. We tested different values for its learning rate (\texttt{lr}). The original value is $6.25e-5$. The best value seems to be a bit higher than the one used to pre-train SP agents ($1e-4$). As expected, very small \texttt{lr} leads to small adaptation. However, a too large value of \texttt{lr} also might result in overshooting and slow down adaptation.

\textbf{Batch size:} Smaller batch size seems to have a consistent improvement in the case of adapting IQL to the OBL partner. However, a more moderate batch size of around 256 seems to work best when adapting an OBL agent to an IQL agent. 

While the best-finetuned agent's adaptation performance still falls short of a few-shot learner's capabilities, it is noteworthy to discover the substantial impact of these HPs on adaptation. However, a more intriguing inquiry is how to construct resilient MARL algorithms that can adapt to a new partner without the requirement of careful HP tuning.

\section{Conclusions and future work}

In this work, we motivate the MARL community to focus on the few-shot adaptation problem besides the well-studied zero-shot coordination problem by giving empirical and intuitive reasons including the failure of current SOTA ZSC algorithms in adapting to new partners. 
Therefore, we propose a benchmark that evaluates the adaptability of MARL algorithms.
Our benchmark targets a realistic scenario where the pre-trained agent is paired with unseen partners that are independently trained with potentially different algorithms or architectures.
We accordingly introduce adaptability metrics that quantify agents' performance during adaptation and perform an empirical study on a diverse of pre-trained Hanabi agents.
Besides, benchmarking several SOTA methods, we performed extensive experiments to investigate the hyper-parameters influencing adaptability. 

We believe that our work paves the way for various promising directions for further investigation in this domain.
Currently, none of the SOTA methods can adapt well to a group of partners in less than thousands of episodes even with careful hyperparameter tuning. Therefore, developing methods that perform well on this benchmark is an important future work for the community.
It would also be valuable to include more recent SOTA methods such as K-level reasoning~\citep{cui2021klevel} to determine their impact on adaptation performance. 
Additionally, investigating the correlation between adaptability and partner diversity would provide insight into the generalizability of the MARL algorithms. Another potential direction for future research is to include other rule-based and human-cloned bots to examine the algorithms' adaptability to different types of partners. Finally, studying the case with learning partners could provide valuable information on the adaptability of MARL algorithms to agents that are continually evolving and learning. Addressing these areas of research could lead to significant advancements in the development of robust and adaptive MARL algorithms.

As one of the limitations of our work, we defined adaptation regret and discussed the choice of partners only for the case of two-player game. However, our benchmark can also be extended to more than the two-player Hanabi game. The definition of adaptation regret remains unchanged, except that $C^*_{j}$ and $C^t_{ij}$ represent the upper-bound performance and the learner $i$'s current performance with a group of partners, rather than with a single partner. However, the process of selecting partners becomes more complex as we now have a multi-dimensional cross-play matrix. For example, how to say in a three-player game, two independent groups of partners of size two are diverse? One natural way would be to evaluate all $2 \times {|P| \choose 3}$ combinations of the partners where $|P|$ is the pool size. Then we can use the same notation of diversity as described before. Although this approach can be extended effortlessly to games with more players, the number of partner combinations grows exponentially with the number of players in the game. Given the popularity of the two-player Hanabi game in the literature, we focus our attention on it in this study. Nevertheless, exploring the setting with more than two players presents an exciting direction for future research.

\section*{Acknowledgements}

This work is supported by the IBM-Mila grant. We acknowledge the computational resources provided by
the Digital Research Alliance of Canada. Janarthanan Rajendran acknowledges the support of the IVADO postdoctoral
fellowship. Sarath Chandar acknowledges the support of the
Canada CIFAR AI Chair program and an NSERC Discovery
Grant.

\bibliography{ref}
\bibliographystyle{collas2023_conference}

\newpage
\appendix

\section{Experimental setup}
\label{sec:experiment-setup}
In this section, we provide the details of our experimental setup including the type of pre-trained agents in the pool, hyper-parameteres used in the experiments, etc.

\subsection{Pre-trained agents types} To create a pool of diverse agents, one way is to per-train agents with different architecture choices. We provide the list of these choices in table~\ref{tab:exact-architectures}.
\begin{table*}[!ht]
\caption{All agent types used in the pool.}
\label{tab:exact-architectures}
\vskip 0.15in
\begin{center}
\begin{small}
\begin{sc}
\begin{tabular}{lcccccccccc}
\toprule
Agent  && RNN type & & Num of feed-forward layers &&  Num of RNN layers && RNN hid dim \\
\midrule
Type-1  && LSTM && 1 && 1 && 256 \\
Type-2  && LSTM && 1 && 1 && 512 \\
Type-3  && LSTM  && 1 && 2 && 256 \\
Type-4  && LSTM &&  1 &&  2 &&  512 \\
Type-5  && LSTM && 2 && 1 && 256 \\
Type-6  && LSTM && 2 && 1 && 512 \\
Type-7  && LSTM && 2 && 2 && 256 \\
Type-8  && LSTM  && 2 && 2 && 512 \\
Type-9  && GRU &&  1 &&  1 &&  256 \\
Type-10  && GRU && 1 && 1 && 512 \\
Type-11  && GRU && 1 && 2 && 256 \\
Type-12 && GRU && 1 && 2 && 512 \\
Type-13  && GRU  && 2 && 1 && 256 \\
Type-14  && GRU &&  2 &&  1 &&  512 \\
Type-15  && GRU && 2 && 2 && 256  \\
Type-16  && GRU && 2 && 2 && 512 \\ 
\bottomrule
\end{tabular}
\end{sc}
\end{small}
\end{center}
\end{table*}

\newpage

\section{Additional results}
\label{sec:additional-exp}

In this section, we discuss the additional results supporting the claims in the main paper. 

\subsection{Finetuning curves of section~\ref{sec:HPS_results}}

We discussed the role of HPs on both adaptivity and final performance of Hanabi agents in section~\ref{sec:HPS_results}. We report the adaptation curves and perfect score curves used to generate Figures~\ref{fig:iql_to_obl_hps} and \ref{fig:obl_to_iql_hps}. 

\begin{figure}[h!]
\centering
\begin{subfigure}[Batchsize - Game score]{
    \centering
    \includegraphics[ width=0.23\linewidth]{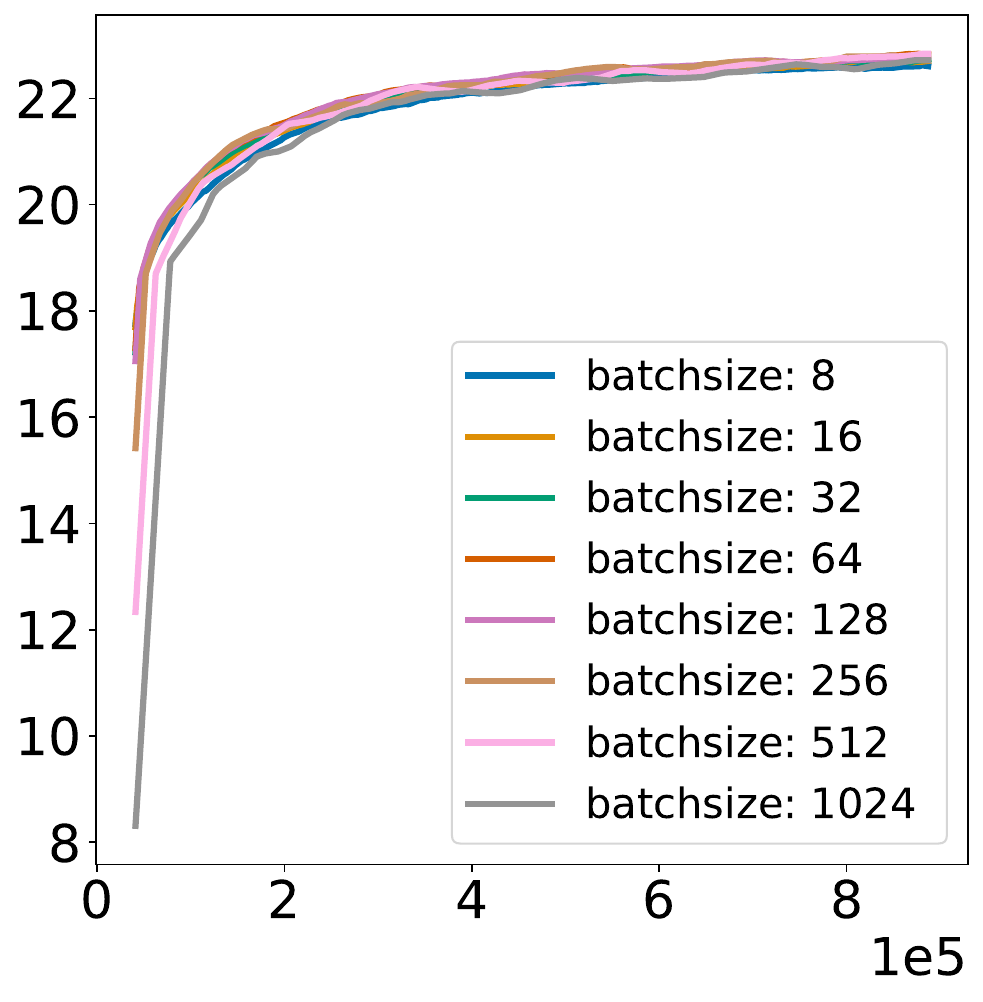}
    }
\end{subfigure}%
\begin{subfigure}[Buffer size - Game score]{
    \centering
    \includegraphics[ width=0.23\linewidth]{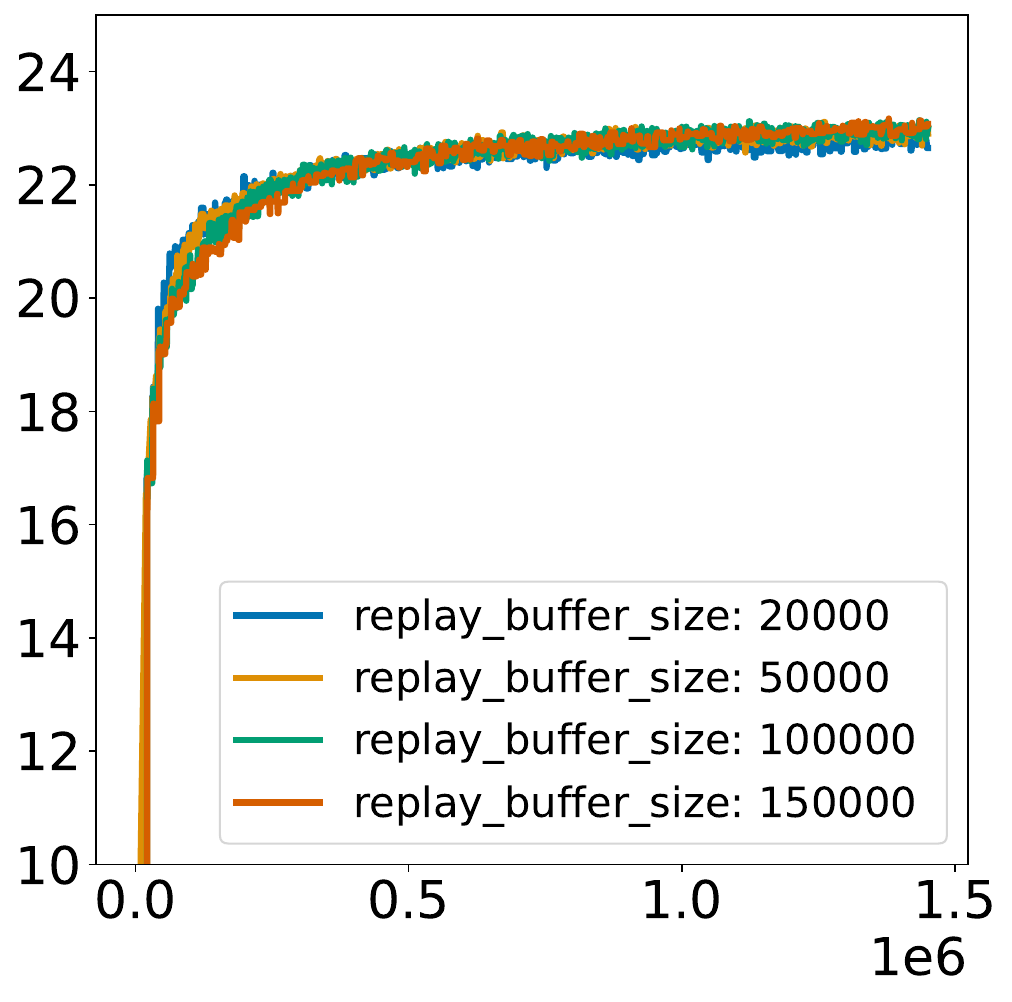}
    }
\end{subfigure}%
\begin{subfigure}[Learning rate - Game score]{
    \centering
    \includegraphics[ width=0.23\linewidth]{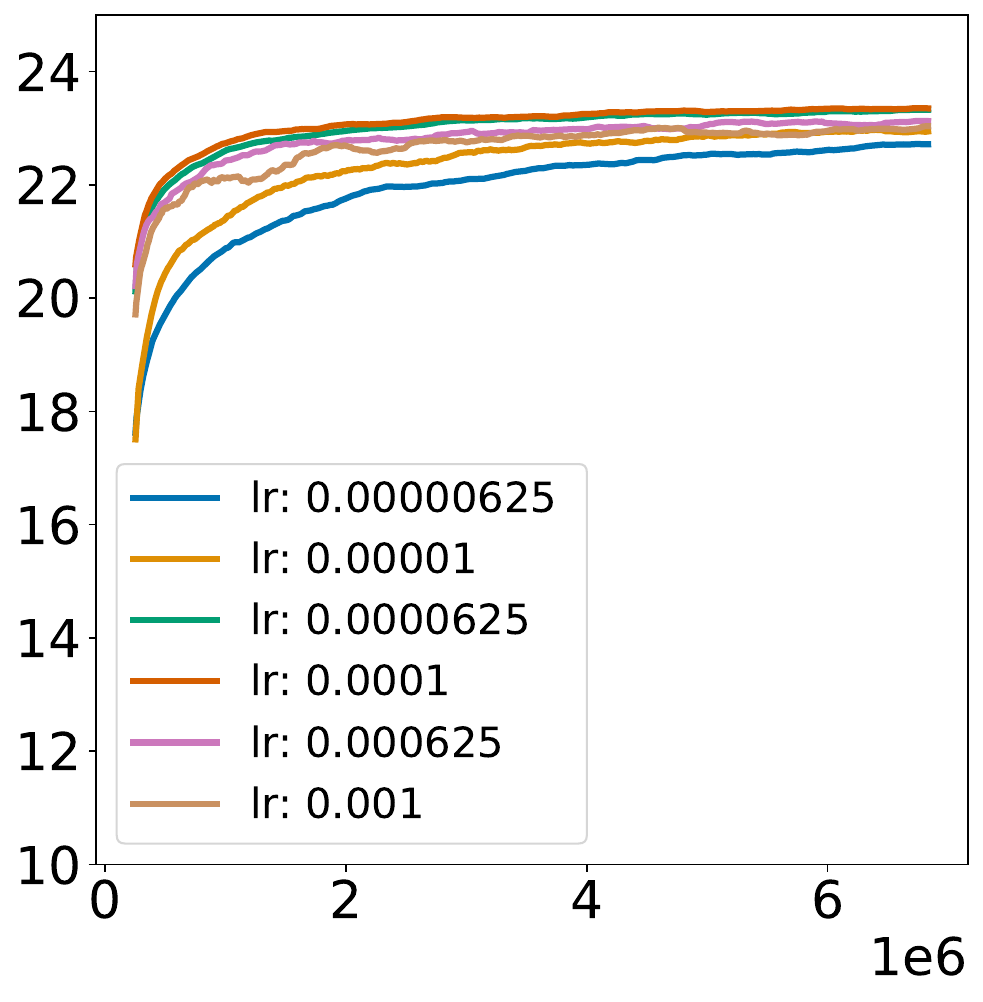}
    }
\end{subfigure}%
\begin{subfigure}[Number of threads - Game score]{
    \centering
    \includegraphics[ width=0.23\linewidth]{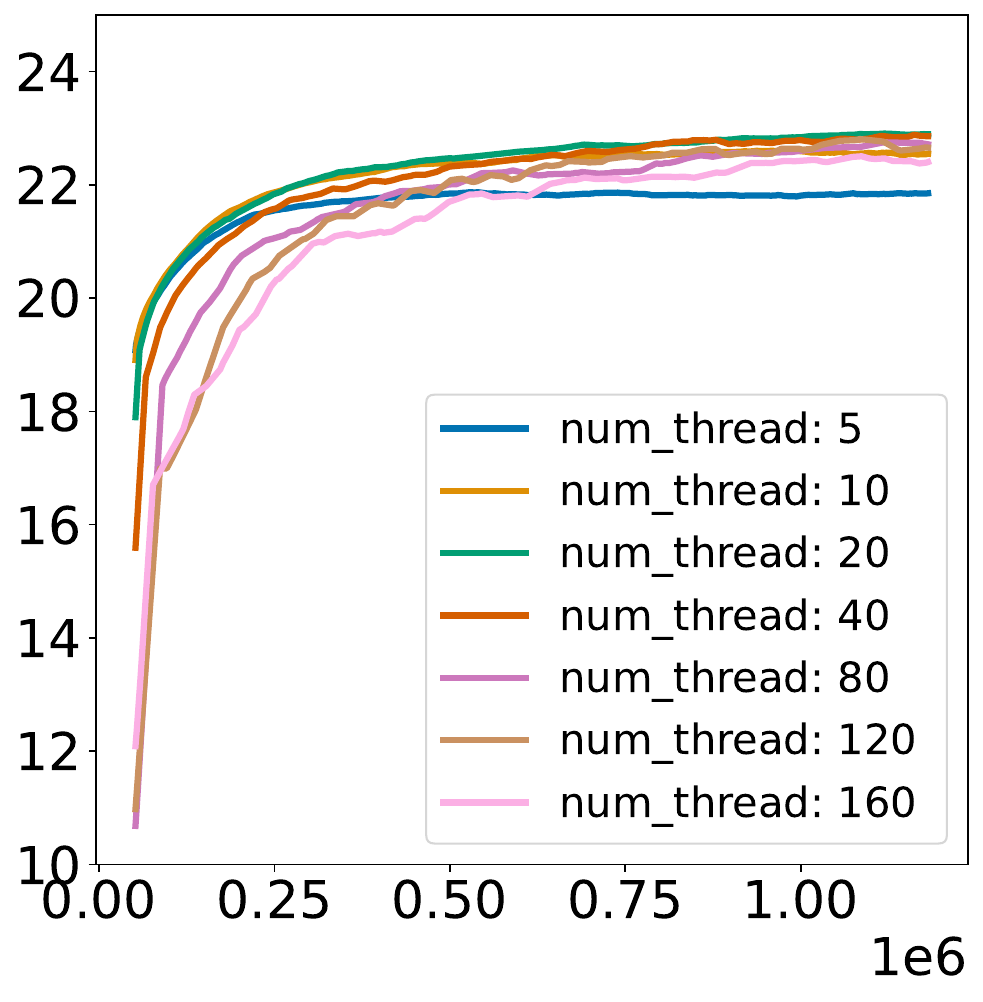}
    }
\end{subfigure}%
\begin{subfigure}[Batchsize - Perfect score]{
    \centering
    \includegraphics[ width=0.23\linewidth]{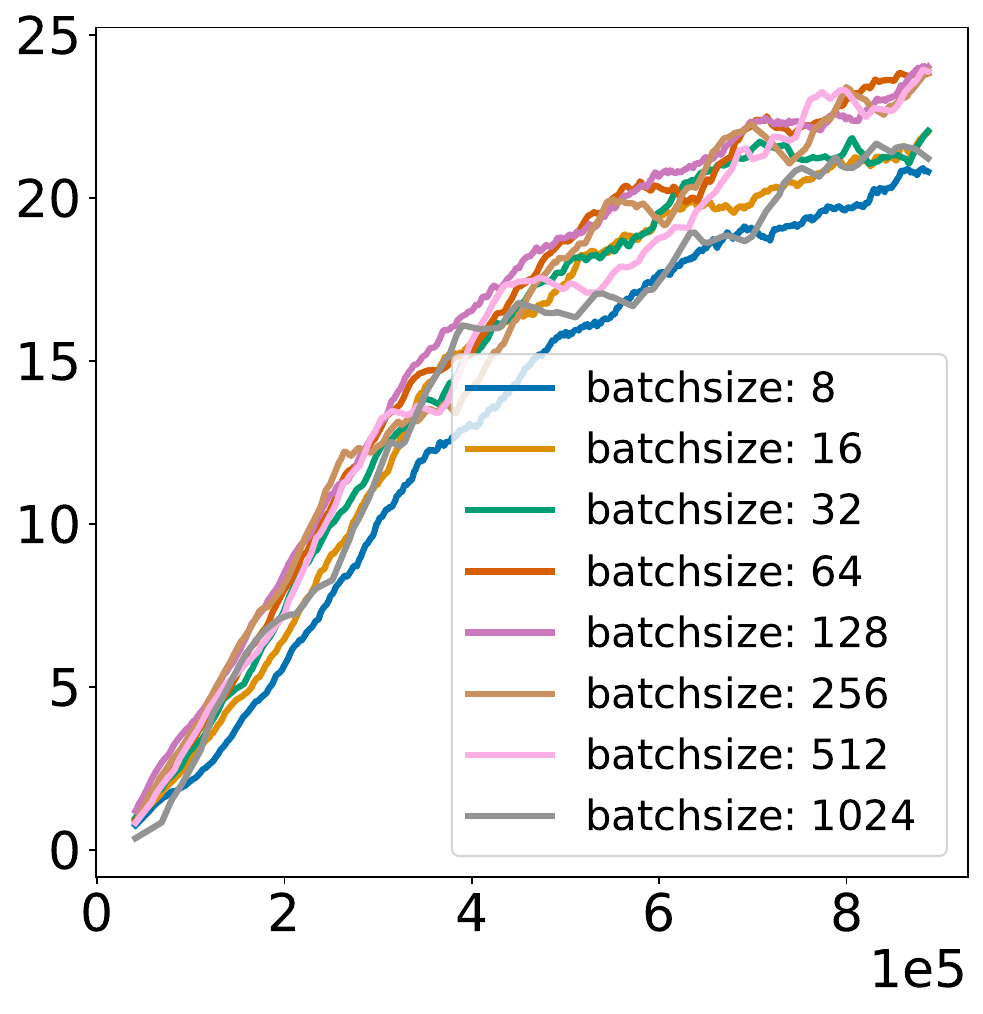}
    }
\end{subfigure}%
\begin{subfigure}[Buffer size - Perfect score]{
    \centering
    \includegraphics[ width=0.23\linewidth]{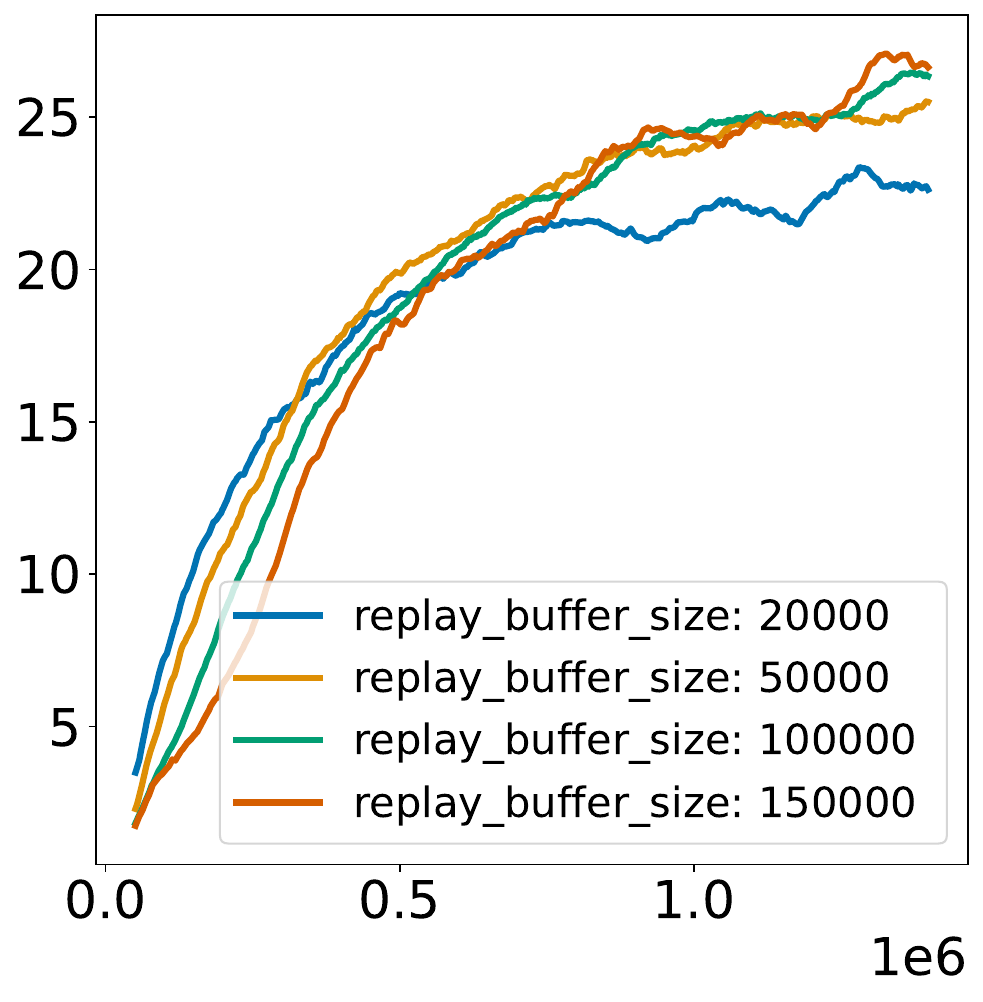}
    }
\end{subfigure}%
\begin{subfigure}[Learning rate - Perfect score]{
    \centering
    \includegraphics[ width=0.23\linewidth]{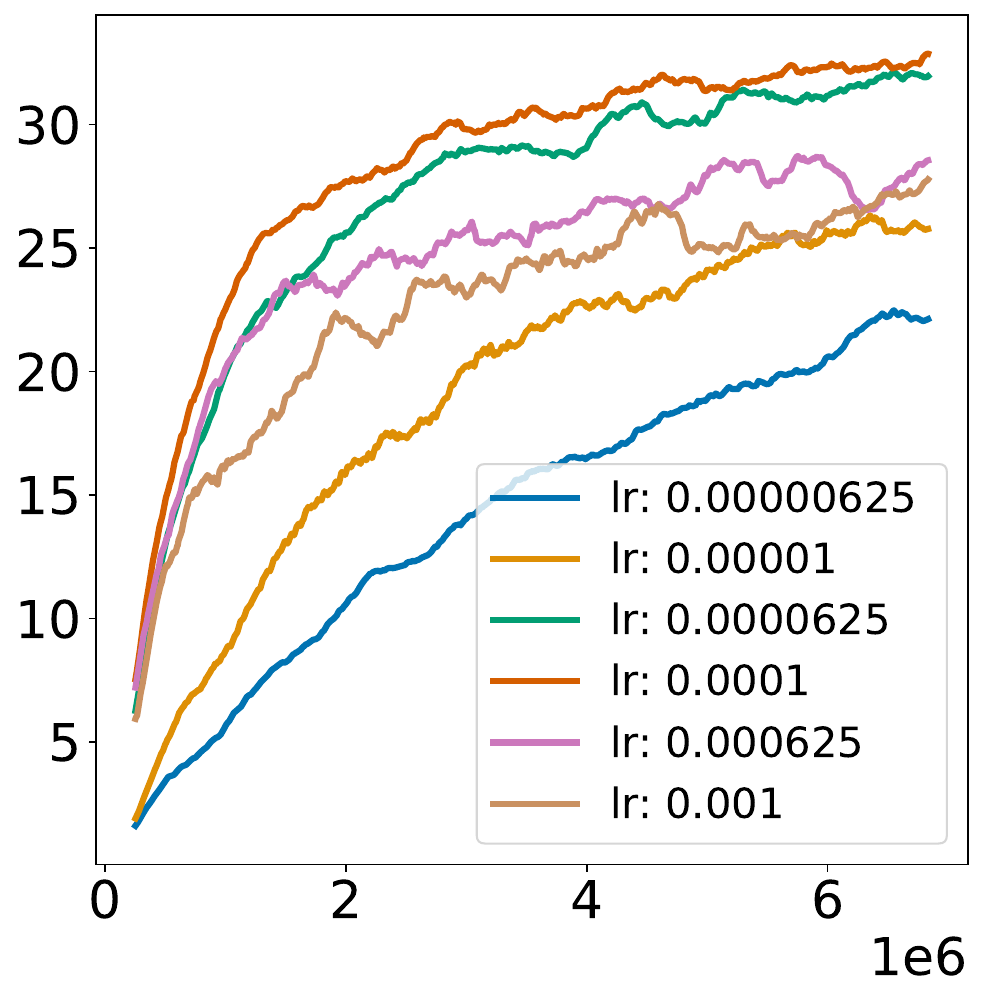}
    }
\end{subfigure}%
\begin{subfigure}[Number of threads - Perfect score]{
    \centering
    \includegraphics[ width=0.23\linewidth]{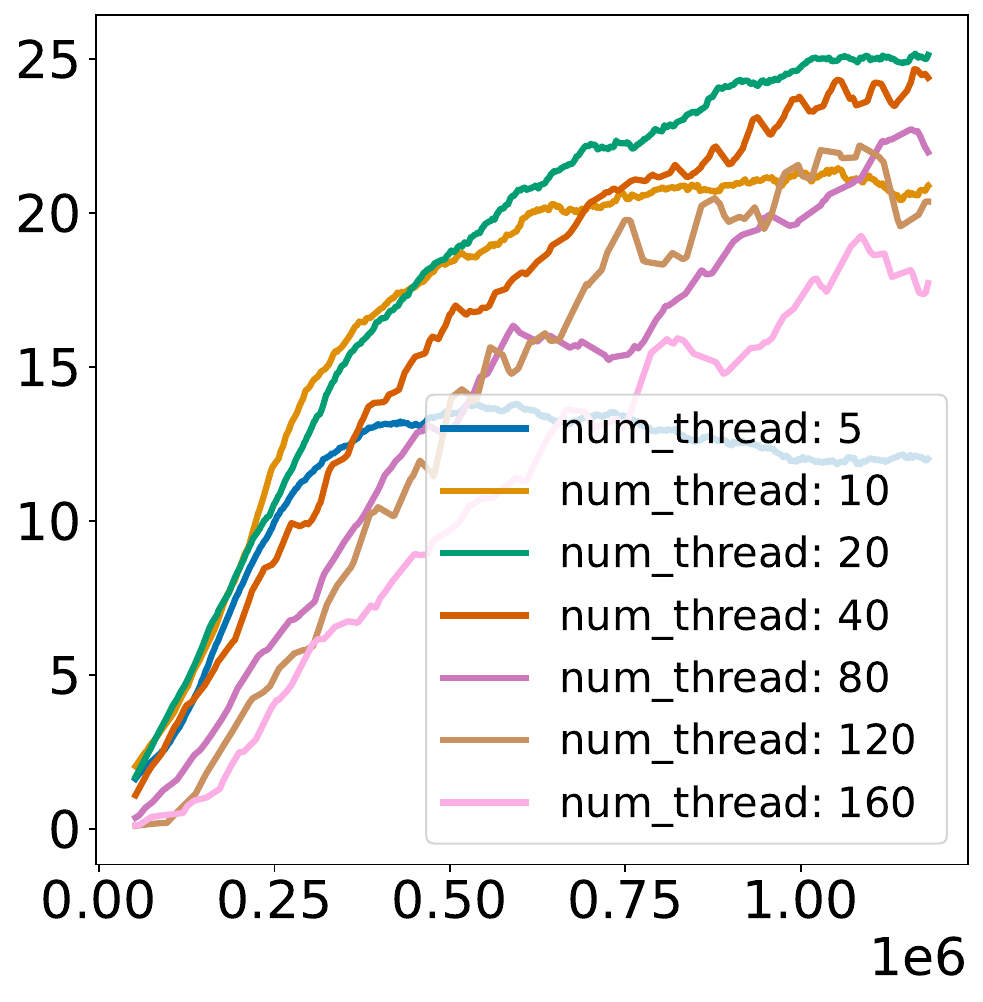}
    }
\end{subfigure}%
\caption{\em The role of different hyperparameters on adaptivity and performance of an OBL agent adapting an IQL agent.}
\label{fig:OBL_IQL_HPs_curves} 
\end{figure}

\newpage
\subsection{The choice of partners}\label{app:choice_of_partners}

\begin{figure}[h!]
\centering

\begin{subfigure}[Total adaptation regret]{
    \centering
    \includegraphics[ width=0.32\linewidth]{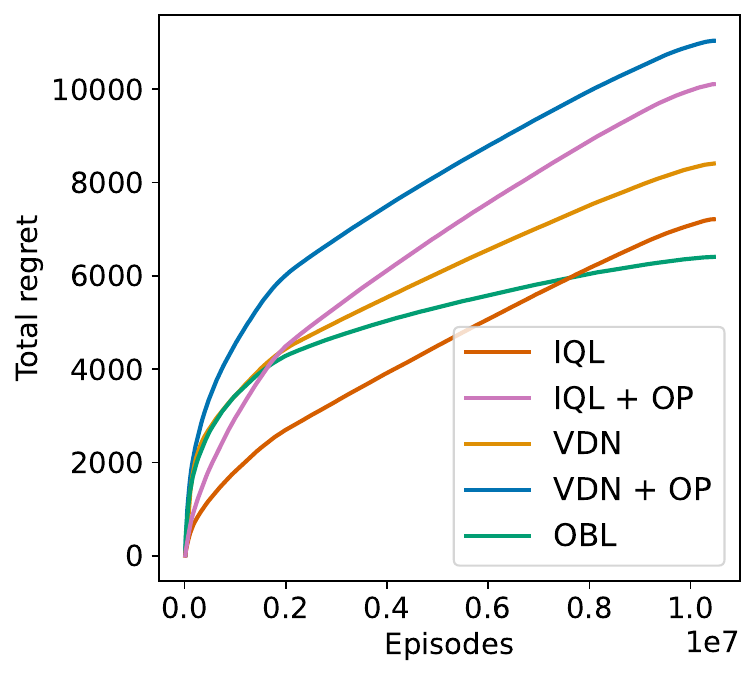}
    }
\end{subfigure}%
\begin{subfigure}[Average adaptation regret]{
    \centering
    \includegraphics[ width=0.3\linewidth]{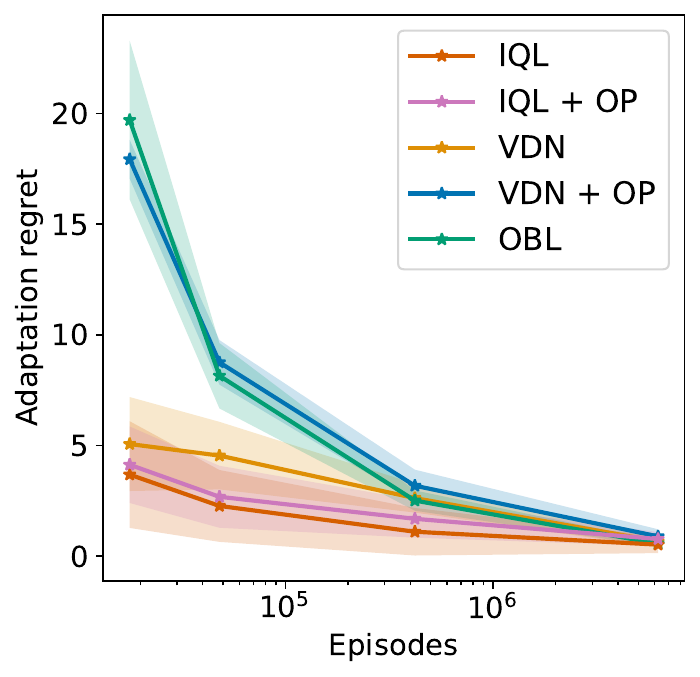}
    }
\end{subfigure}%
\begin{subfigure}[Game score]{
    \centering
    \includegraphics[ width=0.3\linewidth]{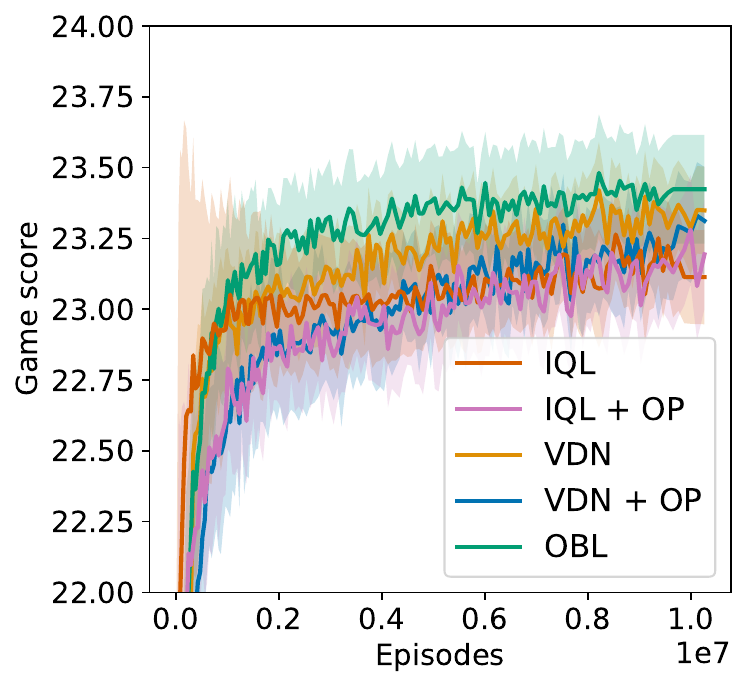}
    }
\end{subfigure}%

\begin{subfigure}[Total adaptation regret]{
    \centering
    \includegraphics[ width=0.32\linewidth]{figs/curves/med_div_part_total_adapt_regret_metric_mean_ub_selfplay.pdf}
    }
\end{subfigure}%
\begin{subfigure}[Average adaptation regret]{
    \centering
    \includegraphics[ width=0.3\linewidth]{figs/curves/med_div_part_mean_adapt_regret_metric_mean_ub_selfplay.pdf}
    }
\end{subfigure}%
\begin{subfigure}[Game score]{
    \centering
    \includegraphics[ width=0.3\linewidth]{figs/curves/med_div_part_score_metric_mean.pdf}
    }
\end{subfigure}%

\begin{subfigure}[Total adaptation regret]{
    \centering
    \includegraphics[ width=0.32\linewidth]{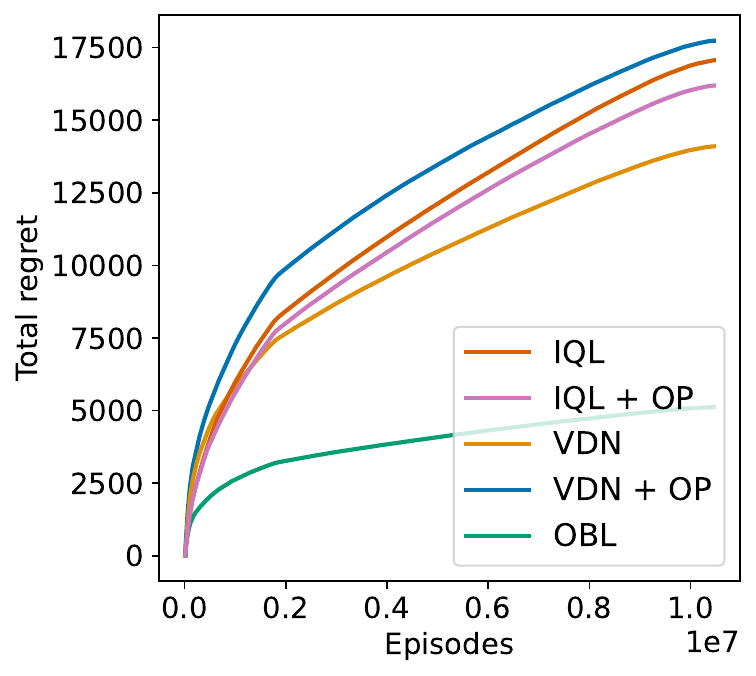}
    }
\end{subfigure}%
\begin{subfigure}[Average adaptation regret]{
    \centering
    \includegraphics[ width=0.3\linewidth]{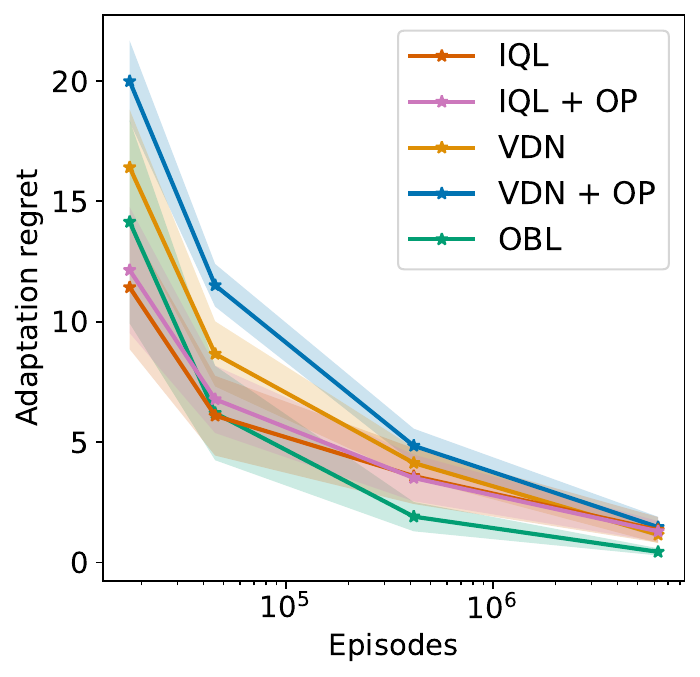}
    }
\end{subfigure}%
\begin{subfigure}[Game score]{
    \centering
    \includegraphics[ width=0.3\linewidth]{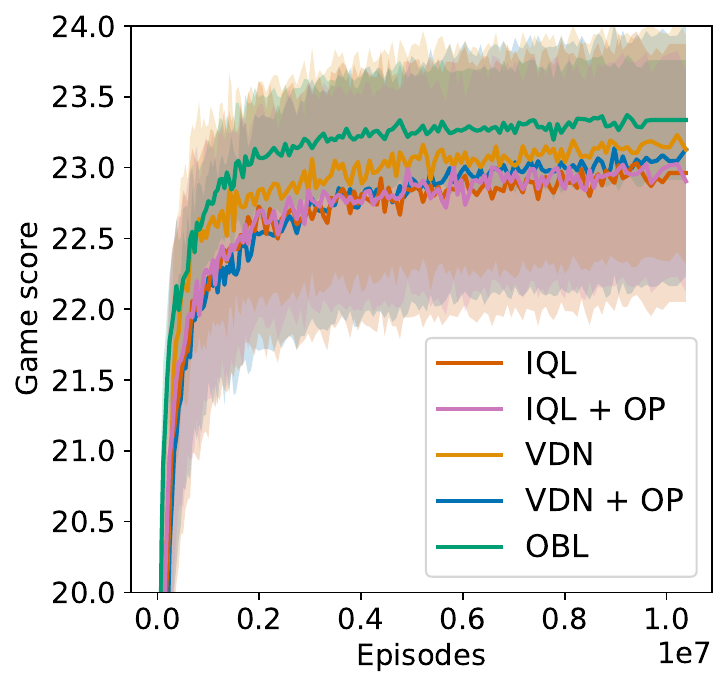}
    }
\end{subfigure}%

\caption{\em Results on adapting 5 different agents to partners from Figure~\ref{fig:partners}(a) (top row) with low diversity level, Figure~\ref{fig:partners}(c) (middle row) with medium diversity, and Figure~\ref{fig:partners}(c) (bottom row) with high divresity. 
}
\label{app_fig:benchmark_results_all_patners} 
\end{figure}

In this section, we report the detailed results for different sets of partners in Figure~\ref{app_fig:benchmark_results_all_patners}. It is clear that the results vary depending on the partners involved. That is why it is crucial to provide all the details about the partner groups when reporting the adaptation results 

\newpage
\subsection{The choice of upper-bound performance}\label{app:upper_bound}

In this section, we analyze the choice of upper-bound performance on the adaptation regret as shown in Figure~\ref{app:upper_bound}. If the SP score of each partner is employed as $C_j^*$, the differences between algorithms become more noticeable regarding the total regret. This is due to the fact that the average regret is already normalized, and modifying the upper-bound score only slightly shifts the regret curves. Nonetheless, this small shift is reflected more prominently in the total regret curve. 

\begin{figure}[h!]
\centering
\begin{subfigure}[Total adaptation regret - $C^*_{j}$ = max game score]{
    \centering
    \includegraphics[ width=0.3\linewidth]{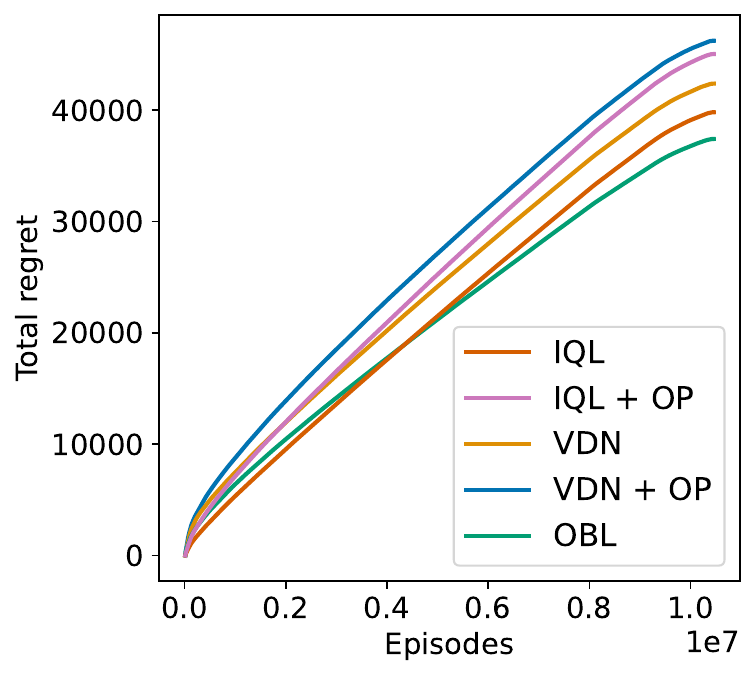}
    }
\end{subfigure}%
\begin{subfigure}[Total adaptation regret - $C^*_{j}$ = Self-Play score]{
    \centering
    \includegraphics[ width=0.3\linewidth]{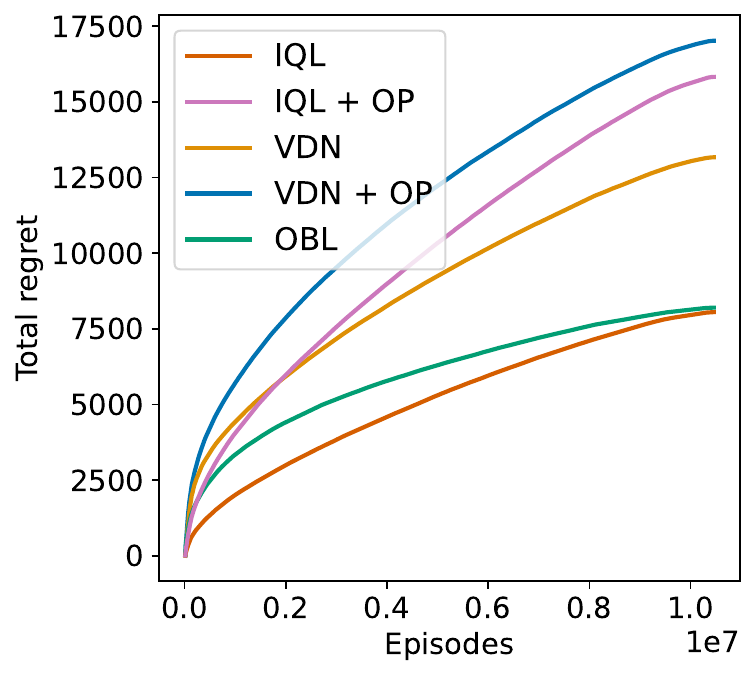}
    }
\end{subfigure} \\
\begin{subfigure}[Average adaptation regret - $C^*_{j}$ = max game score]{
    \centering
    \includegraphics[ width=0.3\linewidth]{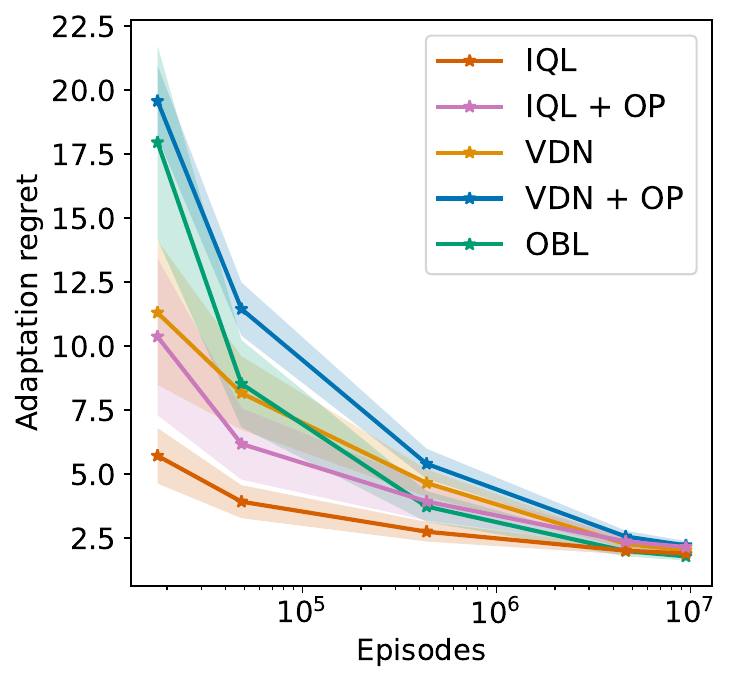}
    }
\end{subfigure}%
\begin{subfigure}[Average adaptation regret - $C^*_{j}$ = Self-Play score]{
    \centering
    \includegraphics[ width=0.3\linewidth]{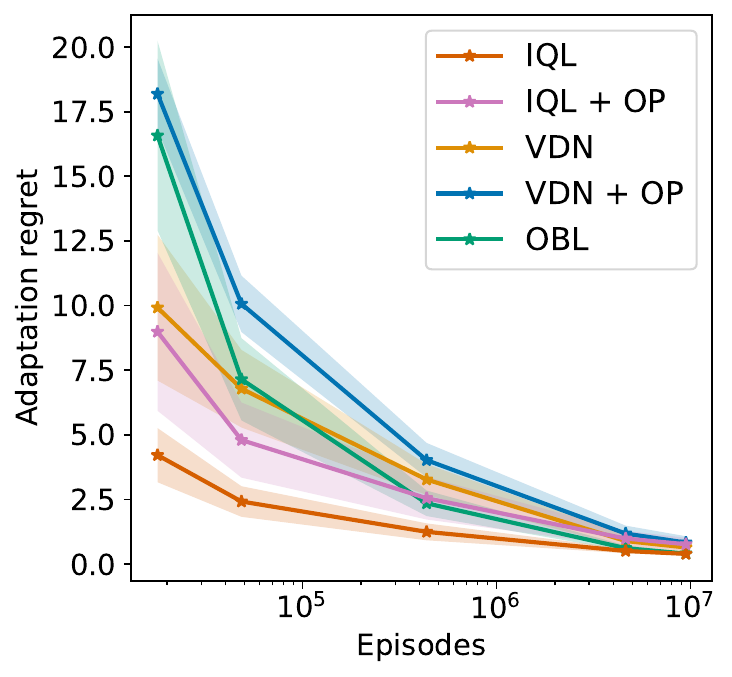}
    }
\end{subfigure}%
\caption{\em The comparison between using different upper-bound performance $C^*_{j}$ in the adaptation regret definition  on benchmark results on adapting 5 different agents to partners from Figure~\ref{fig:partners}(b). It has a negligible
impact on the average adaptation regret. Nevertheless, if the SP score of each partner is
employed as $C^*_{j}$, the differences between algorithms become more noticeable regarding the total regret.}
\label{app:upper_bound} 
\end{figure}

\newpage
\subsection{The choice of the aggregator}\label{app:aggregator}

\begin{figure}[h!]
\centering
\begin{subfigure}[Total adaptation regret - IQM \texttt{aggr}]{
    \centering
    \includegraphics[ width=0.32\linewidth]{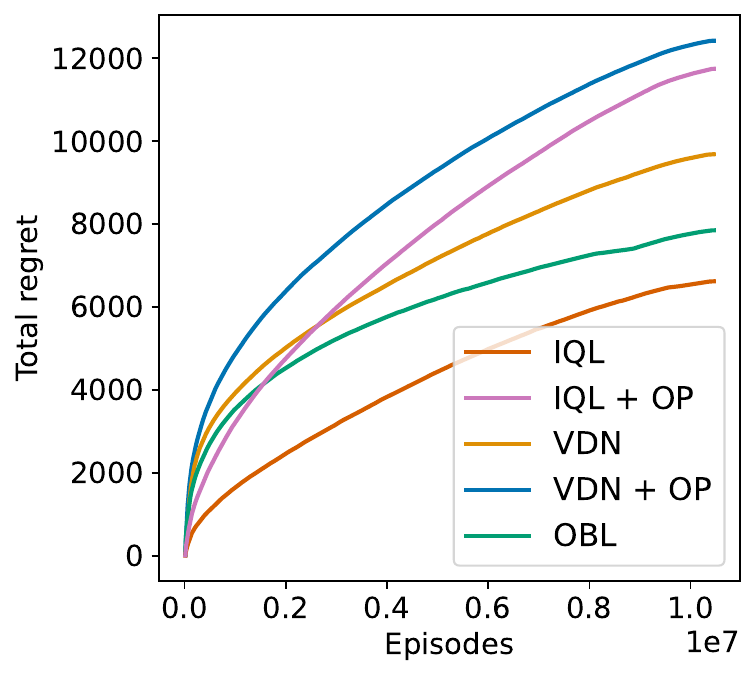}
    }
\end{subfigure}%
\begin{subfigure}[Average adaptation regret - IQM \texttt{aggr}]{
    \centering
    \includegraphics[ width=0.3\linewidth]{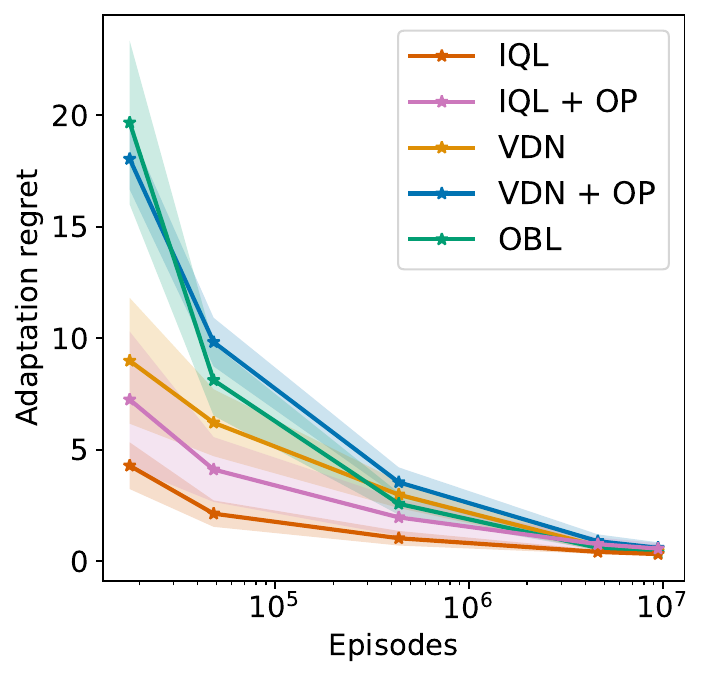}
    }
\end{subfigure}%
\begin{subfigure}[Game score - IQM \texttt{aggr}]{
    \centering
    \includegraphics[ width=0.3\linewidth]{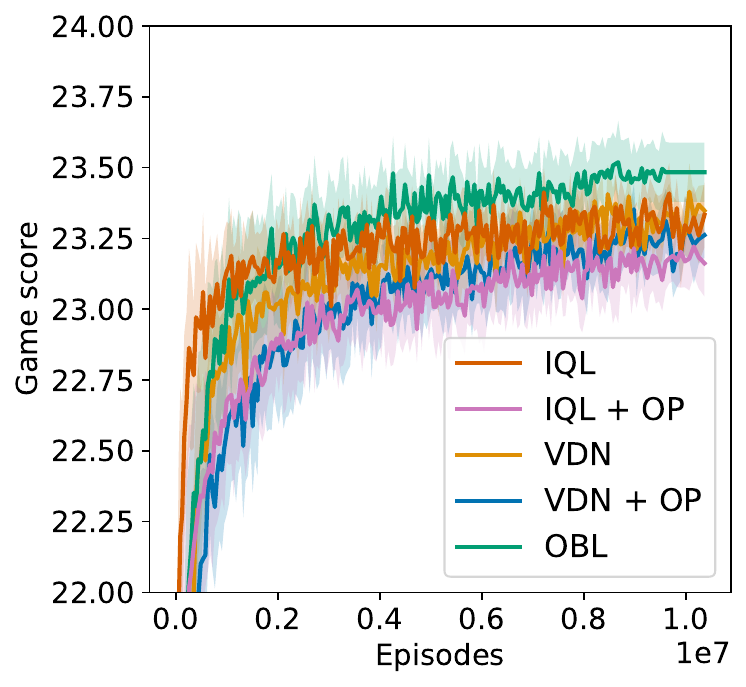}
    }
\end{subfigure}%
\begin{subfigure}[Total adaptation regret - Mean \texttt{aggr}]{
    \centering
    \includegraphics[ width=0.32\linewidth]{figs/total_adapt_regret_metric_mean_ub_selfplay.pdf}
    }
\end{subfigure}%
\begin{subfigure}[Average adaptation regret - Mean \texttt{aggr}]{
    \centering
    \includegraphics[ width=0.3\linewidth]{figs/mean_adapt_regret_metric_mean_ub_selfplay.pdf}
    }
\end{subfigure}%
\begin{subfigure}[Game score- Mean \texttt{aggr}]{
    \centering
    \includegraphics[ width=0.3\linewidth]{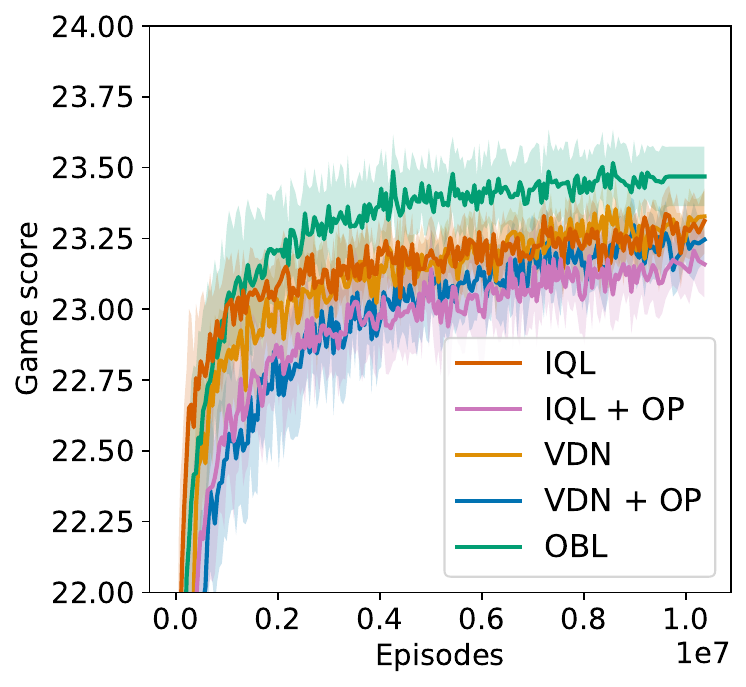}
    }
\end{subfigure}%
\caption{\em The comparison between mean and IQM aggregator on the benchmark results on adapting 5 different agents to partners from Figure~\ref{fig:partners}(b). While the selection of IQM instead of mean aggregator does not alter the ranking in our particular experiments, it is
reflected in the total regret chart by being more resilient to a partner like OBL, which has a considerably low ZSC
performance.
Consequently, the superiority of IQL as the learner over OBL is more evident when employing IQM aggregator.}
\label{fig:benchmark_results_mean_aggr} 
\end{figure}

In this section, we analyze the choice of aggregator on the adaptation regret and game score as shown in Figure~\ref{fig:benchmark_results_mean_aggr}.

\end{document}